\documentclass[pdflatex,sn-mathphys-num]{sn-jnl}

\usepackage{graphicx}%
\usepackage{multirow}%
\usepackage{amsmath,amssymb,amsfonts}%
\usepackage{amssymb}
\usepackage{pifont}
\newcommand{\cmark}{\ding{51}}%
\newcommand{\xmark}{\ding{55}}%
\usepackage{tabularx}

\usepackage{adjustbox}

\usepackage{amsthm}%
\usepackage{mathrsfs}%
\usepackage[title]{appendix}%
\usepackage{xcolor}%
\usepackage{textcomp}%
\usepackage{manyfoot}%
\usepackage{booktabs}%
\usepackage{listings}%
\usepackage{graphicx}
\usepackage{caption}
\usepackage{subcaption} 
\usepackage{booktabs}
\usepackage{amssymb}   
\usepackage[ruled,vlined,linesnumbered]{algorithm2e}

\theoremstyle{thmstyleone}%
\theoremstyle{thmstyletwo}%

\theoremstyle{thmstylethree}%

\begin{document}

\title[Article Title]{MultiSHAP: A Shapley-Based Framework for Explaining Cross-Modal Interactions in Multimodal AI Models}


\author[1,3]{\fnm{Zhanliang} \sur{Wang}}\email{aaronwzl@sas.upenn.edu}

\author*[1,2,3]{\fnm{Kai} \sur{Wang}}\email{wangk@chop.edu}

\affil[1]{\orgdiv{Department of Mathematics}, \orgname{University of Pennsylvania}, \orgaddress{\street{209 S 33rd St}, \city{Philadelphia}, \postcode{19104}, \state{PA}, \country{United States}}}

\affil[2]{\orgdiv{Department of Pathology and Laboratory Medicine}, \orgname{University of Pennsylvania}, \orgaddress{\street{3400 Spruce Street}, \city{Philadelphia}, \postcode{19104}, \state{PA}, \country{United States}}}

\affil*[3]{\orgdiv{Raymond G. Perelman Center for Cellular and Molecular Therapeutics}, \orgname{Children's Hospital of Philadelphia}, \orgaddress{\street{3501 Civic Center Blvd}, \city{Philadelphia}, \postcode{19104}, \state{PA}, \country{United States}}}

\abstract{
Multimodal AI models have achieved impressive performance in tasks that require integrating information from multiple modalities, such as vision and language. However, their ``black-box'' nature poses a major barrier to deployment in high-stakes applications where interpretability and trustworthiness are essential. How to explain cross-modal interactions in multimodal AI models remains a major challenge. While existing model explanation methods, such as attention map and Grad-CAM, offer coarse insights into cross-modal relationships, they cannot precisely quantify the synergistic effects between modalities, and are limited to open-source models with accessible internal weights. Here we introduce \emph{MultiSHAP}, a model-agnostic interpretability framework that leverages the \emph{Shapley Interaction Index} to attribute multimodal predictions to \emph{pairwise} interactions between fine-grained visual and textual elements (such as image patches and text tokens), while being applicable to both open- and closed-source models. Our approach provides: $(1)$ instance-level explanations that reveal synergistic and suppressive cross-modal effects for individual samples - ``\emph{why the model makes a specific prediction on this input}'', and $(2)$ dataset-level explanation that uncovers generalizable interaction patterns across samples - ``\emph{how the model integrates information across modalities}''. Experiments on public multimodal benchmarks confirm that MultiSHAP faithfully captures cross-modal reasoning mechanisms, while real-world case studies demonstrate its practical utility. Our framework is extensible beyond two modalities, offering a general solution for interpreting complex multimodal AI models.}




\maketitle

\section*{Introduction}
\label{sec:intro}

Multimodal artificial intelligence (AI) systems have achieved remarkable performance on tasks requiring the integration of vision and language, including visual question answering (VQA)~\cite{antol2015vqa, goyal2017making} and image-text retrieval~\cite{lin2014microsoft, young2014image}. Models such as CLIP~\cite{radford2021learningtransferablevisualmodels}, ViLT~\cite{kim2021viltvisionandlanguagetransformerconvolution}, and LLaVA~\cite{liu2023improvedllava} align image patches with text tokens to form joint representations for semantic understanding. Although these models yield accurate predictions, their internal decision processes, particularly how specific visual and textual elements interact remain poorly understood.

This opacity is especially concerning in high-stakes domains such as medical AI, where interpretability is essential for safe deployment~\cite{rodis2024multimodalexplainableartificialintelligence,xiao2025restoringcalibrationalignedlarge, huang2022interpretability}. In rare disease diagnosis, for instance, models must integrate phenotype descriptions with patient images to support clinical decision-making~\cite{wu2025mint, wu2025integratingchainofthoughtretrievalaugmented, 10.1145/3765612.3767763}. Understanding which features from each modality contribute to a diagnosis~\cite{hou2025fairccafairrepresentation} and how they interact is vital for establishing trust, identifying failure modes, and guiding model improvements. However, existing explainability techniques such as Grad-CAM~\cite{Selvaraju_2019} or attention maps~\cite{attentionmap} offer only coarse visualizations and cannot quantify whether interactions between specific patches and tokens are supportive or misleading. Furthermore, these gradient-based methods require access to internal network layers, rendering them unsuitable for interpreting closed-source models.
\begin{figure*}[h]
    \centering
    \includegraphics[width=\textwidth]{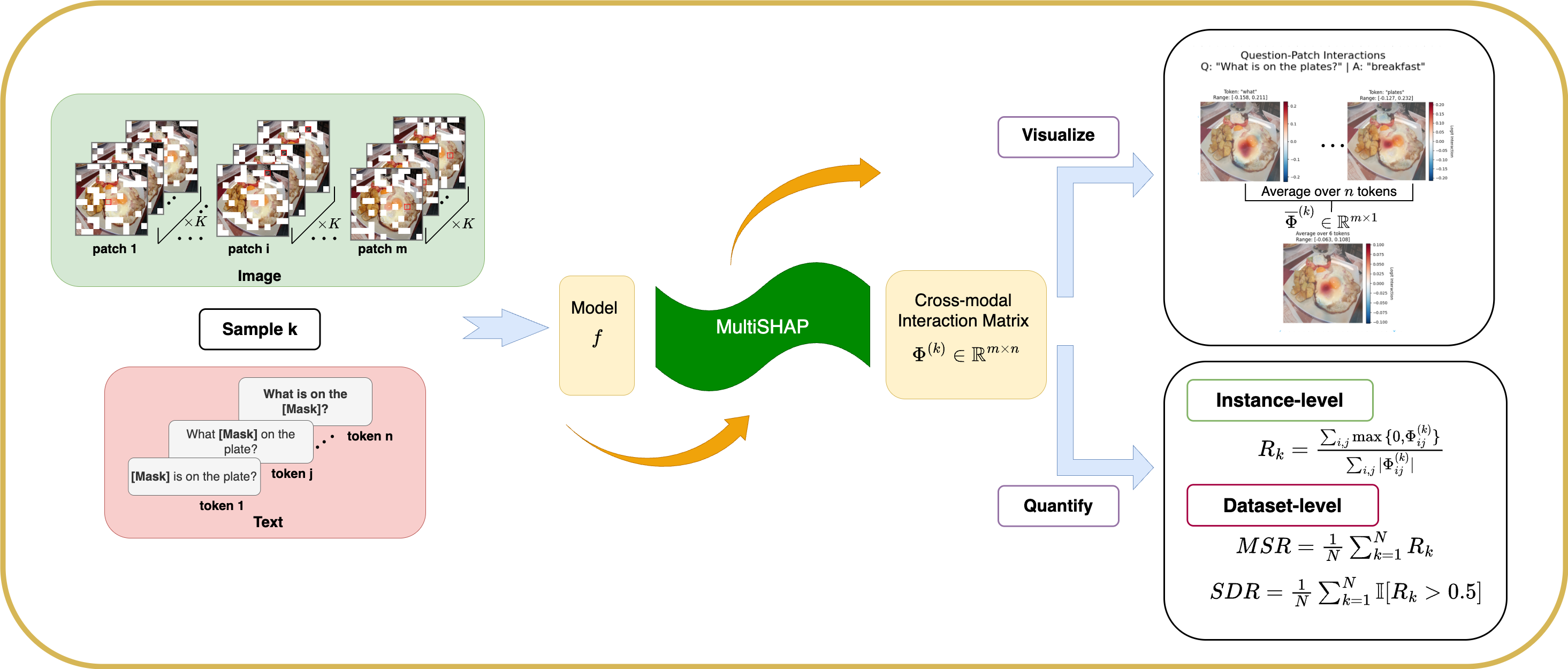}
   \caption{\textbf{Overview of the MultiSHAP workflow.}
For a sample $k$, the input image is partitioned into $m$ patches and the text query into $n$ tokens. The model $f$ is evaluated on masked patch--token combinations, and MultiSHAP estimates a cross-modal interaction matrix $\boldsymbol{\Phi}^{(k)}\in\mathbb{R}^{m\times n}$, where $\Phi^{(k)}_{ij}$ denotes the Shapley interaction between image patch $i$ and text token $j$. Interactions are approximated via Monte Carlo sampling with $K$ coalitions per sample. The resulting matrix can be \emph{visualized} as token-specific interaction heatmaps and aggregated cross-modal attribution maps (e.g., averaged over tokens). It can also be \emph{quantified} using interaction-based interpretability metrics: the synergy ratio $R_k$ summarizes, for each instance, the relative dominance of synergistic (positive) versus suppressive (negative) interactions; at the dataset level, the Mean Synergy Ratio (MSR) measures the average tendency toward synergistic interactions, and the Synergy Dominance Ratio (SDR) reports the fraction of samples in which synergy outweighs suppression. Positive (red) and negative (blue) values indicate synergistic and suppressive cross-modal interactions, respectively.}

    \label{fig:multishap_workflow}
\end{figure*}
Here we present MultiSHAP, a general and model-agnostic framework for interpreting multimodal predictions by quantifying fine-grained cross-modal interactions (Fig.~\ref{fig:multishap_workflow}). MultiSHAP leverages the Shapley Interaction Index from cooperative game theory to compute the synergistic (positive) or suppressive (negative) effect of each image patch and text token pair on the model output. By systematically masking combinations of visual and textual elements, our method estimates how their joint presence impacts predictions beyond their individual contributions. This yields an interpretable interaction matrix that reveals how image and text elements collaborate or conflict during inference.

Unlike existing attribution methods that provide only modality-level explanations (InterSHAP~\cite{Wenderoth_2025}) or unimodal attributions (TokenSHAP~\cite{goldshmidt2024tokenshapinterpretinglargelanguage}, PixelSHAP~\cite{goldshmidt2025attentionpleasepixelshapreveals}), MultiSHAP explicitly quantifies patch-token interactions, enabling identification of localized synergistic and suppressive effects that drive multimodal decisions. In contrast to attention-based techniques, which are architecture-dependent and often conflate correlation with causation, MultiSHAP provides faithful, counterfactual explanations grounded in axiomatic game-theoretic principles.

MultiSHAP supports both instance-level and dataset-level analysis through interpretable metrics that summarize interaction strength and patterns. We apply MultiSHAP to VQA and image-text retrieval tasks, evaluating performance on standard benchmarks (VQAv2, MSCOCO, Flickr30K) and a medical dataset (GestaltMatcher Database) for rare disease diagnosis. Our experiments reveal four distinct cross-modal interaction patterns: beneficial synergy supporting correct predictions, harmful synergy causing errors, helpful suppression filtering misleading cues, and detrimental suppression undermining accuracy. These findings demonstrate MultiSHAP's potential for improving explainability and trust in multimodal AI applications across scientific domains. 

\section*{Methods}
\label{sec:methods}

\subsection*{Problem formulation and notation}

We formulate multimodal interpretability as quantifying how image patches and text tokens interact to influence model predictions. Our approach extends recent work on Shapley-based modality attribution~\cite{mmshap2023, Wenderoth_2025} to fine-grained patch-token interactions, yielding a cross-modal interaction matrix $\boldsymbol{\Phi}$. The framework is model-agnostic, requiring only the ability to query the model with masked inputs. Without loss of generality, we describe the method using image and text as input modalities.

A multimodal sample is denoted $X=(\mathcal{I},\mathcal{T})$, where $\mathcal{I}\in\mathbb{R}^{H\times W\times C}$ is an input image and $\mathcal{T}=\{t_1,\dots,t_n\}$ is a sequence of $n$ tokenized text elements. The image is partitioned into $m=\frac{HW}{s^{2}}$ non-overlapping patches of size $s \times s$: $\mathcal{P}=\{p_1,\dots,p_m\}\subset\mathbb{R}^{d_v}$, where each patch $p_i$ has visual feature dimension $d_v$. The combined feature set is $\mathcal{M}=\mathcal{P}\cup\mathcal{T}$ with $|\mathcal{M}|=m+n$ total features.

For any subset $S\subseteq\mathcal{M}$ and model $f$, we define the aggregated representations:
\begin{align}
z_v(S) &= f_v(S\cap\mathcal{P}) \in\mathbb{R}^{d} \quad \text{(visual embedding)}\\
z_t(S) &= f_t(S\cap\mathcal{T}) \in\mathbb{R}^{d} \quad \text{(textual embedding)}
\end{align}
The model outputs a scalar score via cross-modal fusion: $v(S)=g\bigl(z_v(S),z_t(S)\bigr)\in\mathbb{R}$, where $g: \mathbb{R}^d \times \mathbb{R}^d \to \mathbb{R}$ represents the multimodal scoring function.

\subsection*{Task-specific score functions}

The score function $v(S)$ is task-dependent. For visual question answering, we use the logit for the predicted answer class:
\begin{equation}
v(S) = f(\text{mask}(X, S))_{y^*}
\end{equation}
where $y^*$ is the ground truth answer class. For image-text retrieval, we use cosine similarity between visual and textual embeddings:
\begin{equation}
v(S) = \frac{z_v(S) \cdot z_t(S)}{\|z_v(S)\| \cdot \|z_t(S)\|}
\end{equation}

\subsection*{Shapley Interaction Index}

To quantify how individual patches and tokens interact synergistically or suppressively, we leverage the Shapley Interaction Index from cooperative game theory~\cite{shapley1953value, lundberg2017unified, tsai2022faith}. This captures second-order effects beyond individual feature contributions. For each patch-token pair $(p_i,t_j)$, the interaction strength is defined as:
\begin{equation}
\label{eq:SII}
\Phi_{ij}=
\sum_{S\subseteq\mathcal{M}\setminus\{p_i,t_j\}}
\frac{|S|!\,(|\mathcal{M}|-|S|-2)!}{2(|\mathcal{M}|-1)!}\;
\Delta_{ij}(S),
\end{equation}
where the discrete second-order difference measures the joint contribution:
\begin{align}
\label{eq:delta}
\Delta_{ij}(S) &= v(S\cup\{p_i,t_j\}) - v(S\cup\{p_i\}) - v(S\cup\{t_j\}) + v(S)
\end{align}

The resulting interaction matrix $\boldsymbol{\Phi}\in\mathbb{R}^{m\times n}$ captures synergistic interactions ($\Phi_{ij}>0$), where the patch-token pair contributes more together than the sum of their individual contributions, and suppressive interactions ($\Phi_{ij}<0$), where the joint presence reduces the combined contribution, indicating conflict or redundancy.

\subsection*{Monte Carlo approximation}

Since exact computation requires $O(2^{m+n-2})$ model evaluations, we use Monte Carlo sampling~\cite{montecarlo2022}. We randomly sample $K$ coalitions $\{S_k\}_{k=1}^{K}$ and estimate:
\begin{align}
\hat{\Phi}_{ij}= \frac{1}{K}\sum_{k=1}^{K}\Bigl[
v(S_k\cup\{p_i,t_j\})-v(S_k\cup\{p_i\})-v(S_k\cup\{t_j\})+v(S_k)\Bigr]
\end{align}

We employ stratified sampling over coalition sizes to reduce estimation variance. In practice, $K = 32$--$128$ samples provide stable estimates while maintaining computational efficiency with $O(K \times m \times n)$ model evaluations.

\subsection*{Algorithm implementation}

Algorithm~\ref{alg:MultiSHAP} presents the complete MultiSHAP procedure for estimating the cross-modal interaction matrix. The algorithm implements Monte Carlo estimation through four key stages: coalition sampling, input masking, interaction computation, and result aggregation.

\begin{algorithm}[t]
\caption{MultiSHAP: Estimating Cross-Modal Interaction Matrix $\boldsymbol{\Phi}$}
\label{alg:MultiSHAP}

\KwIn{Image patches $\mathcal{P} = \{p_1, \ldots, p_m\}$, text tokens $\mathcal{T} = \{t_1, \ldots, t_n\}$, model $f$, masking function $\text{mask}(\cdot, \cdot)$, number of samples $K$}
\KwOut{Cross-modal interaction matrix $\boldsymbol{\Phi} \in \mathbb{R}^{m \times n}$}

$\boldsymbol{\Phi} \leftarrow \mathbf{0}_{m \times n}$;\quad
$\mathbf{W} \leftarrow \mathbf{0}_{m \times n}$\;

\For{$k \leftarrow 1$ \KwTo $K$}{
    Sample coalition $\mathcal{S} \subseteq \{1, \ldots, m+n\}$ uniformly at random\;
    $v_{\mathcal{S}} \leftarrow f(\text{mask}(\mathcal{P} \cup \mathcal{T}, \mathcal{S}))$\;

    \For{$i \leftarrow 1$ \KwTo $m$}{
        \For{$j \leftarrow m+1$ \KwTo $m+n$}{
            \If{$i \notin \mathcal{S} \land j \notin \mathcal{S}$}{
                $v_{\mathcal{S} \cup \{i,j\}} \leftarrow f(\text{mask}(\mathcal{P} \cup \mathcal{T}, \mathcal{S} \cup \{i,j\}))$\;
                $v_{\mathcal{S} \cup \{i\}} \leftarrow f(\text{mask}(\mathcal{P} \cup \mathcal{T}, \mathcal{S} \cup \{i\}))$\;
                $v_{\mathcal{S} \cup \{j\}} \leftarrow f(\text{mask}(\mathcal{P} \cup \mathcal{T}, \mathcal{S} \cup \{j\}))$\;

                $\Delta_{i,j-m} \leftarrow v_{\mathcal{S} \cup \{i,j\}} - v_{\mathcal{S} \cup \{i\}} - v_{\mathcal{S} \cup \{j\}} + v_{\mathcal{S}}$\;
                $\boldsymbol{\Phi}_{i,j-m} \leftarrow \boldsymbol{\Phi}_{i,j-m} + \Delta_{i,j-m}$\;
                $\mathbf{W}_{i,j-m} \leftarrow \mathbf{W}_{i,j-m} + 1$\;
            }
        }
    }
}

$\boldsymbol{\Phi} \leftarrow \boldsymbol{\Phi} \oslash \mathbf{W}$\;
\Return $\boldsymbol{\Phi}$\;

\end{algorithm}

For each coalition $\mathcal{S}$, we compute $\Delta_{ij}(\mathcal{S})$ for all absent patch-token pairs $(p_i, t_j)$ to measure their joint contribution. The masking function creates inputs as follows:
\begin{align}
\text{mask}(\mathcal{I}, S_v) &= \begin{cases} 
p_i & \text{if } i \in S_v \\
\mathbf{0} & \text{if } i \notin S_v 
\end{cases} \quad \forall i \in \{1, \ldots, m\} \\
\text{mask}(\mathcal{T}, S_t) &= \begin{cases} 
t_j & \text{if } j \in S_t \\
\texttt{[MASK]} & \text{if } j \notin S_t 
\end{cases} \quad \forall j \in \{1, \ldots, n\}
\end{align}
where $S_v = S \cap \{1, \ldots, m\}$ and $S_t = S \cap \{m+1, \ldots, m+n\}$ represent the visual and textual feature subsets. This masking strategy preserves the input structure required by the multimodal model while systematically ablating specific features.

\subsection*{Interpretability metrics}

We define comprehensive metrics to characterize interaction patterns at both instance and dataset levels. For each sample $k$ with interaction matrix $\boldsymbol{\Phi}^{(k)}\in\mathbb{R}^{m\times n}$, we compute:
\begin{align}
T_k &=\textstyle\sum_{i,j}\!|\Phi^{(k)}_{ij}| &&\text{(total interaction strength)}\\
S_k &=\sum_{i,j}\!\max\{0,\Phi^{(k)}_{ij}\} &&\text{(synergy strength)}\\
P_k &=\sum_{i,j}\!\max\{0,-\Phi^{(k)}_{ij}\} &&\text{(suppression strength)}\\
R_k &=S_k/T_k\in[0,1] &&\text{(synergy ratio)}
\end{align}

The synergy ratio $R_k$ serves as a key indicator: high values ($R_k > 0.5$) suggest the model relies primarily on collaborative cross-modal processing, while low values indicate suppression-dominated reasoning.

At the dataset level, for dataset $\mathcal{D}=\{(x_k,y_k)\}_{k=1}^{N}$, we compute:
\begin{align}
\text{MSR} &= \frac{1}{N}\sum_{k=1}^{N} R_k \quad \text{(Mean Synergy Ratio)}\\
\text{SDR} &= \frac{1}{N}\sum_{k=1}^{N}\mathbb{I}[R_k > 0.5] \quad \text{(Synergy Dominance Ratio)}
\end{align}

Mean Synergy Ratio (MSR) measures the average tendency toward synergistic interactions across the dataset. Synergy Dominance Ratio (SDR) quantifies the proportion of samples where synergistic interactions outweigh suppressive ones. Together, these metrics enable systematic analysis of model behaviour patterns and comparison across architectures and domains.

\subsection*{Visualization}

To facilitate interpretation, we provide multiple visualization modes (Fig.~\ref{fig:multishap_workflow}): token-wise heatmaps showing interactions between specific tokens and image regions, and aggregated spatial maps displaying average interaction patterns across all tokens. These visualizations enable both fine-grained analysis of specific cross-modal relationships and high-level understanding of model attention patterns.

\subsection*{Datasets}

We evaluated MultiSHAP on four benchmarks across two multimodal tasks:

\textbf{Visual Question Answering.} VQAv2~\cite{goyal2017making} is a general-domain benchmark containing natural images with questions requiring visual reasoning. GestaltMatcher Database (GMDB)~\cite{gmdb2022} is a medical dataset for rare disease diagnosis from facial photographs, containing images of patients with genetic syndromes paired with diagnostic questions.

\textbf{Image-Text Retrieval.} MSCOCO~\cite{lin2014microsoft} contains natural images with literal descriptive captions. Flickr30K~\cite{flickr30k} contains images with more compositional and varied caption styles.

\subsection*{Models}

For VQA tasks, we used ViLT-VQA~\cite{kim2021viltvisionandlanguagetransformerconvolution} on VQAv2 and GestaltMML~\cite{gestaltmml2023} on GMDB. Both models use $224 \times 224$ input resolution with $32 \times 32$ patches, yielding $m = 49$ image patches. For image-text retrieval, we used fine-tuned CLIP ViT-B/32~\cite{radford2021learningtransferablevisualmodels} on both MSCOCO and Flickr30K with the same resolution and patch configuration. All models were fine-tuned on their respective datasets to ensure strong baseline performance before interpretability analysis.

\subsection*{Experimental setup}

All experiments were conducted on a MacBook Pro equipped with an Apple M2 Max chip and 32GB of RAM. For each dataset, we randomly sampled 500 samples and report results averaged over 3 random seeds to ensure robustness. To estimate Shapley interaction scores, we applied Monte Carlo sampling with $K=128$ permutations per sample following standard practice.

\subsection*{Computational complexity}

MultiSHAP requires $O(K \times m \times n)$ model evaluations, where $K$ is the number of Monte Carlo samples. With $K=128$, this is significantly more efficient than exact Shapley Interaction Index computation, which requires $O(2^{m+n})$ evaluations. The stratified sampling strategy reduces the required $K$ by approximately 30\% compared to uniform sampling while maintaining estimation quality.

\subsection*{Runtime analysis}

Runtime scales roughly linearly with the number of Monte Carlo sampled coalitions $K$: on an Apple M2 Max, MultiSHAP takes 17.5s per sample at $K=32$, 37.2s at $K=68$, and $70.0$s at $K=128$ (Table~\ref{tab:runtime}).

\subsection*{Robustness Analysis}

All quantitative results are reported as mean $\pm$ standard deviation across 3 random seeds. For dataset-level metrics (MSR, SDR), we computed statistics over 500 randomly sampled examples per dataset.


\section*{Results}
\label{sec:results}

\subsection*{MultiSHAP reveals four distinct cross-modal interaction patterns}
\begin{table*}[htbp!]
\centering

\begin{adjustbox}{max width=\textwidth}
\begin{tabular}{llllccccc}
\toprule
\textbf{Task} & \textbf{Dataset} & \textbf{ID} & \textbf{Pred.} &
$\mathbf{T_k}$ & $\mathbf{S_k}$ & $\mathbf{P_k}$ & $\mathbf{R_k}$ & \textbf{Type} \\
\midrule
VQA       & GMDB      & Ex.~$1$  & \cmark  & $84.51$ & $45.59$ & $38.92$ & $0.5394$ & Synergistic \\
VQA       & GMDB      & Ex.~$2$  & \xmark  & $67.78$ & $23.36$ & $27.41$ & $0.4601$ & Suppressive \\
VQA       & VQAv2     & Ex.~$3$  & \cmark  & $83.45$ & $47.23$ & $36.22$ & $0.5652$ & Synergistic \\
VQA       & VQAv2     & Ex.~$4$  & \cmark  & $79.38$ & $32.74$ & $46.64$ & $0.4084$ & Suppressive \\
VQA       & VQAv2     & Ex.~$5$  & \xmark  & $74.73$ & $46.48$ & $28.25$ & $0.6219$ & Synergistic \\
VQA       & VQAv2     & Ex.~$6$  & \xmark  & $67.65$ & $22.21$ & $30.87$ & $0.4188$ & Suppressive \\
Retrieval & MSCOCO    & Ex.~$7$  & GT      & $96.43$ & $55.05$ & $41.38$ & $0.5709$ & Synergistic \\
Retrieval & MSCOCO    & Ex.~$8$  & Foil    & $88.05$ & $41.74$ & $46.31$ & $0.4741$ & Suppressive \\
Retrieval & Flickr30K & Ex.~$9$  & GT      & $63.66$ & $38.01$ & $25.65$ & $0.5970$ & Synergistic \\
Retrieval & Flickr30K & Ex.~$10$ & Foil    & $66.09$ & $32.93$ & $34.06$ & $0.4982$ & Suppressive \\
\bottomrule
\end{tabular}
\end{adjustbox}

\vspace{2mm}
\raggedright
\small
\caption{\textbf{Sample-level MultiSHAP statistics for representative cases.}
ID: sample index; Pred.: prediction (\cmark=Correct, \xmark=Incorrect, GT=Ground Truth, Foil=semantic-similar distractor to the ground truth caption.);
$T_k$: total interaction strength; $S_k$: synergy strength; $P_k$: suppression strength; $R_k$: synergy ratio; Type: interaction type.}
\label{tab:sample_summary}
\end{table*}
We evaluated MultiSHAP on two multimodal tasks: visual question answering (VQA) using VQAv2~\cite{goyal2017making} and GestaltMatcher Database (GMDB)~\cite{gmdb2022}, and image-text retrieval using MSCOCO~\cite{lin2014microsoft} and Flickr30K~\cite{flickr30k}. For each sample, MultiSHAP computes an interaction matrix $\boldsymbol{\Phi} \in \mathbb{R}^{m \times n}$ quantifying synergistic ($\Phi_{ij} > 0$) and suppressive ($\Phi_{ij} < 0$) interactions between image patches and text tokens (see Methods). Analysis of representative cases (Table~\ref{tab:sample_summary}) reveals four distinct interaction patterns that characterize multimodal reasoning behavior.

\subsection*{Synergistic interactions enable accurate medical diagnosis}

\begin{figure*}[!htbp]
    \centering

    \begin{subfigure}[t]{0.24\textwidth}
        \centering
        \includegraphics[width=\textwidth]{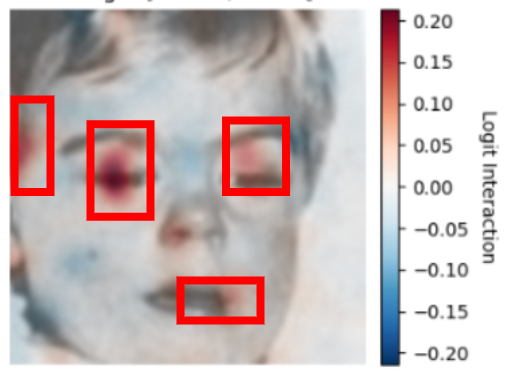}
        \caption{\textbf{Example 1\\ (Correct)}\\
       \textit{Q:} What is the most likely diagnosis?\\
        \footnotesize\textit{A:} CdLS}
        \label{fig:main_examples_a}
    \end{subfigure}
    \hfill
    \begin{subfigure}[t]{0.24\textwidth}
        \centering
        \includegraphics[width=\textwidth]{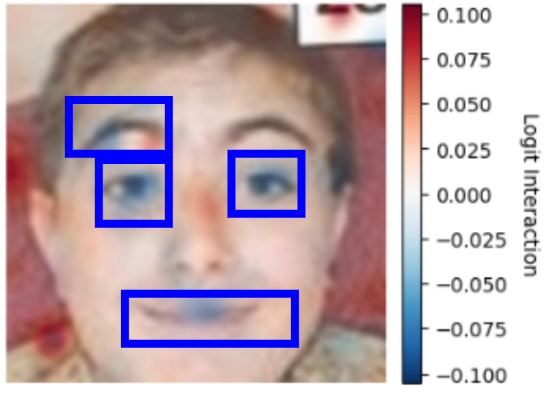}
        \caption{\textbf{Example 2\\ (Incorrect)}\\
        \footnotesize\textit{Q:} What is the most likely diagnosis?\\
        \footnotesize\textit{A:} Robinow syndrome}
        \label{fig:main_examples_b}
    \end{subfigure}
    \hfill
    \begin{subfigure}[t]{0.24\textwidth}
        \centering
        \includegraphics[width=\textwidth]{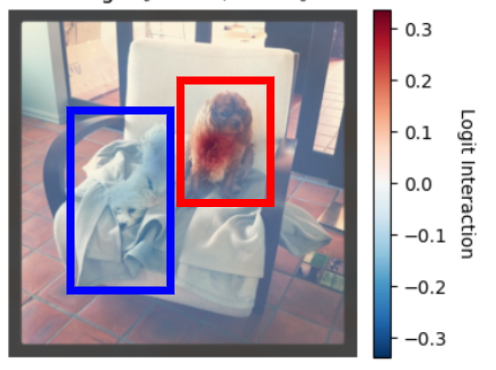}
        \caption{\textbf{Example 4\\ (Correct)}\\
        \footnotesize\textit{Q:} Are both dogs white?\\
        \footnotesize\textit{A:} No}
        \label{fig:main_examples_c}
    \end{subfigure}
    \hfill
    \begin{subfigure}[t]{0.24\textwidth}
        \centering
        \includegraphics[width=\textwidth]{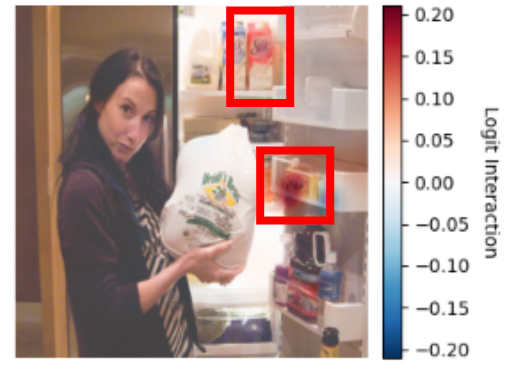}
        \caption{\textbf{Example 5\\ (Incorrect)}\\
        \footnotesize\textit{Q:} What color is the top of the bottle?\\
        \footnotesize\textit{A:} Orange (should be white)}
        \label{fig:main_examples_d}
    \end{subfigure}

    {\footnotesize\textbf{\textcolor{red}{Red} = synergistic (positive) interaction; \textcolor{blue}{Blue} = suppressive (negative) interaction.}}

    \caption{\textbf{MultiSHAP reveals distinct cross-modal interaction patterns.}
    Each heatmap visualizes patch--token interactions for one sample. Synergistic interactions (positive) highlight evidence that mutually reinforces across modalities, whereas suppressive interactions (negative) indicate conflicting evidence. In \textbf{(a)}, synergy concentrates on diagnostically relevant facial regions and yields a correct rare-disease prediction. In \textbf{(b)}, interactions emphasize less informative regions, corresponding to an incorrect diagnosis. In \textbf{(c)}, suppression helps downweight misleading visual cues and supports a correct VQA answer, while in \textbf{(d)} spurious synergy with irrelevant objects contributes to failure. See Appendix~\ref{app:token_heatmaps} for additional examples (Example 3 and Example 6) and Supplementary Information for token-wise analysis.}
    \label{fig:main_examples}
\end{figure*}

We first examined MultiSHAP's ability to interpret medical image analysis using the GMDB dataset for rare disease diagnosis. Figure~\ref{fig:main_examples}a illustrates a correct diagnosis of Cornelia de Lange Syndrome (CdLS) by the GestaltMML model. CdLS is characterized by distinctive facial features including synophrys (joined eyebrows), long philtrum, and depressed nasal bridge~\cite{Deardorff2020CdLS}. The MultiSHAP heatmap reveals strong synergistic interactions (red regions) between the diagnostic question and clinically relevant facial areas such as glabella, eyes, and philtrum, corresponding to known CdLS phenotypic markers. The synergy-dominated interaction pattern ($S_k = 45.59$, $P_k = 38.92$, $R_k = 0.5394$) indicates effective cross-modal integration supporting accurate clinical decision-making.


\subsection*{Inappropriate suppression causes diagnostic errors}

In contrast, Figure~\ref{fig:main_examples}b shows a misdiagnosis where the model incorrectly predicts Robinow syndrome for a CdLS patient presenting with hypertelorism (increased distance between eyes) and a prominent mouth. Despite phenotypic similarity between syndromes, MultiSHAP reveals predominant suppressive interactions (blue regions) in diagnostically important eye and mouth regions. The low synergy ratio ($R_k = 0.4601$) reflects poor cross-modal alignment where critical visual evidence is inappropriately down-weighted, leading to diagnostic error. This demonstrates how MultiSHAP can identify failure modes by revealing when important visual evidence is suppressed rather than integrated with textual queries.

\subsection*{Suppressive interactions can benefit visual reasoning}

Not all suppressive interactions are detrimental. Figure~\ref{fig:main_examples}c presents a VQAv2 example where the model correctly answers ``No'' to the question ``Are both dogs white?'' Despite suppression-dominated interactions ($P_k = 46.64$ vs.\ $S_k = 32.74$, $R_k = 0.4084$), the model succeeds because suppressive interactions serve a beneficial role. While the brown dog shows strong positive interactions supporting the negative answer, suppressive interactions with the white dog help disambiguate by reducing potentially misleading evidence that might support a ``Yes'' response. This illustrates that suppressive interactions can function as a filtering mechanism for irrelevant or contradictory visual cues.

\begin{figure}[!htbp]
    \centering

    \begin{subfigure}[t]{0.24\textwidth}
        \centering
        \includegraphics[width=\textwidth]{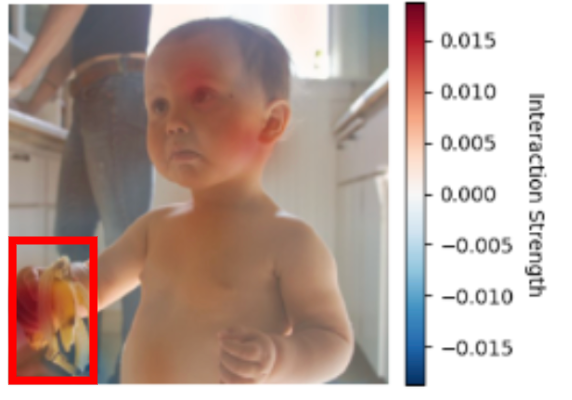}
        \caption{\textbf{Example 7\\ (MSCOCO, GT)}\\
        \footnotesize\textit{Caption:} A baby holding a banana in his right hand.}
        \label{fig:retrieval_a}
    \end{subfigure}
    \hfill
    \begin{subfigure}[t]{0.24\textwidth}
        \centering
        \includegraphics[width=\textwidth]{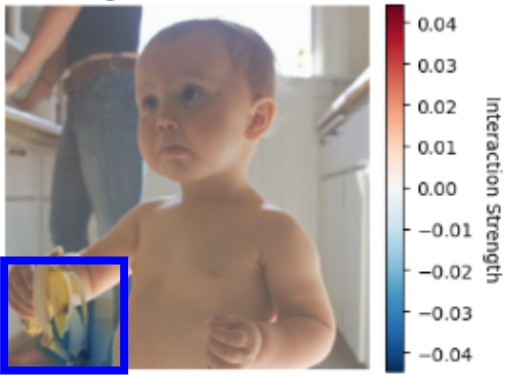}
        \caption{\textbf{Example 8\\ (MSCOCO, Foil)}\\
        \footnotesize\textit{Caption:} A baby holding a watermelon in his left hand.}
        \label{fig:retrieval_b}
    \end{subfigure}
    \hfill
    \begin{subfigure}[t]{0.24\textwidth}
        \centering
        \includegraphics[width=\textwidth]{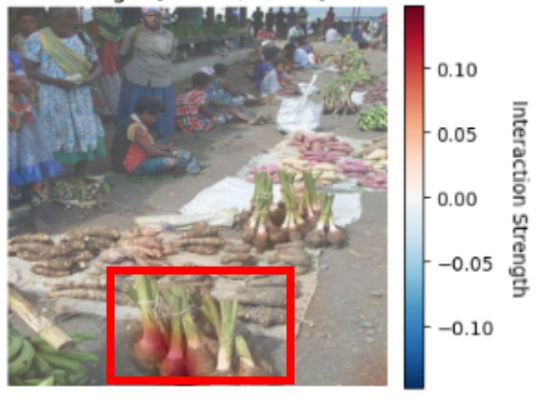}
        \caption{\textbf{Example 9\\ (Flickr30K, GT)}\\
        \footnotesize\textit{Caption:} There are some very large onions.}
        \label{fig:retrieval_c}
    \end{subfigure}
    \hfill
    \begin{subfigure}[t]{0.24\textwidth}
        \centering
        \includegraphics[width=\textwidth]{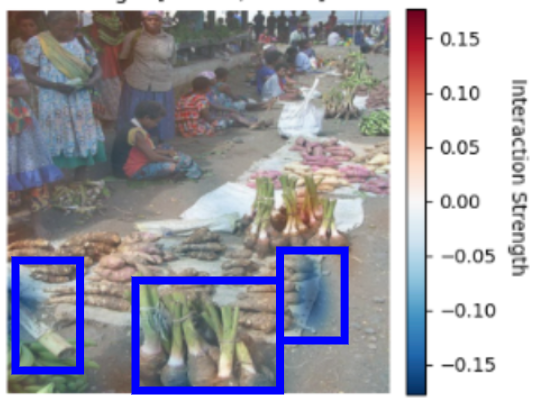}
        \caption{\textbf{Example 10\\ (Flickr30K, Foil)}\\
        \footnotesize\textit{Caption:} There are some very large watermelons.}
        \label{fig:retrieval_d}
    \end{subfigure}

    {\footnotesize\textbf{\textcolor{red}{Red} = synergistic (positive) interaction; \textcolor{blue}{Blue} = suppressive (negative) interaction.}}

    \caption{\textbf{MultiSHAP captures semantic alignment in image--text retrieval.}
    Each panel contrasts the interaction patterns induced by a ground-truth (GT) caption versus a foil caption describing a mismatched object. 
    In \textbf{(a)} and \textbf{(c)}, GT captions yield concentrated synergistic interactions on the correct visual evidence (e.g., banana, onions).
    In \textbf{(b)} and \textbf{(d)}, foil captions induce suppressive interactions over the true object regions, indicating semantic mismatch.
    Together, these examples illustrate how MultiSHAP differentiates aligned versus misaligned image--text pairs through patch--token interactions.}
    \label{fig:retrieval_examples}
\end{figure}

\subsection*{Spurious synergy leads to prediction failures}

Figure~\ref{fig:main_examples}d shows a failure case where the model incorrectly answers ``orange'' instead of ``white'' to the question ``What color is the top of the bottle?'' Despite some correct interactions with the white bottle cap, strong synergistic interactions ($R_k = 0.6219$) with irrelevant colorful objects in the lower refrigerator area cause the model to predict incorrectly. Token-wise analysis reveals that spatial tokens such as ``top'' fail to focus attention appropriately, allowing visually dominant but semantically incorrect cues to influence reasoning. This demonstrates how misaligned positive interactions can amplify irrelevant visual evidence, leading to erroneous conclusions even when synergy dominates the interaction pattern.

\subsection*{MultiSHAP captures semantic alignment in image-text retrieval}

We extended our analysis to image-text retrieval tasks using MSCOCO and Flickr30K (Fig.~\ref{fig:retrieval_examples}). For MSCOCO, the ground-truth caption ``A baby holding a banana in his right hand'' (Example 7) shows strong synergistic interactions ($R_k = 0.5709$) concentrated on the banana region, indicating effective visual-textual grounding. In contrast, the semantically similar foil caption ``A baby holding a watermelon in his left hand'' (Example 8) exhibits suppressive interactions ($R_k = 0.4741$) over the actual banana region, demonstrating the model's ability to detect object substitutions and spatial mismatches.

Flickr30K examples reveal consistent patterns: the ground-truth caption ``There are some very large onions'' (Example 9) exhibits focused positive interactions ($R_k = 0.5970$) with the onion regions, while the foil ``There are some very large watermelons'' (Example 10) triggers suppressive responses ($R_k = 0.4982$) in the same regions. This shows how the model appropriately down-weights visual evidence that contradicts textual descriptions, effectively filtering hallucinated object references. These retrieval patterns confirm that MultiSHAP successfully captures both positive semantic alignment and negative mismatch detection across different multimodal architectures.

\subsection*{Dataset-level analysis reveals task-specific interaction patterns}

Beyond instance-level analysis, MultiSHAP provides dataset-level metrics that characterize overall model behavior (Table~\ref{tab:main_metrics}). We computed Mean Synergy Ratio (MSR) and Synergy Dominance Ratio (SDR) across 500 randomly sampled examples per dataset (see Methods for metric definitions).

\begin{table}[htbp!]
\centering
\small
\setlength{\tabcolsep}{6pt}
\begin{tabular}{llccc}
\toprule
\textbf{Task} & \textbf{Dataset} & \textbf{Accuracy} & \textbf{MSR} & \textbf{SDR} \\
\midrule
\multirow{2}{*}{VQA (ViLT)} 
    & VQAv2     & $0.7456 \pm 0.0339$ & $0.5152 \pm 0.0052$ & $0.5293 \pm 0.0338$ \\
    & GMDB      & $0.6274 \pm 0.0324$ & $0.5168 \pm 0.0104$ & $0.5314 \pm 0.0081$ \\
\midrule
\multirow{2}{*}{Retrieval (CLIP)} 
    & MSCOCO    & ---                 & $0.5583 \pm 0.0217$ & $0.5084 \pm 0.0989$ \\
    & Flickr30K & ---                 & $0.5367 \pm 0.0125$ & $0.5633 \pm 0.0125$ \\
\bottomrule
\end{tabular}

\vspace{2mm}
\raggedright
\small
\caption{\textbf{Dataset-level performance and MultiSHAP interaction metrics.} Accuracy is reported for VQA tasks only. MSR: Mean Synergy Ratio; SDR: Synergy Dominance Ratio. Values are mean $\pm$ s.d.\ across 3 random seeds with 500 samples each.}
\label{tab:main_metrics}
\end{table}

Interestingly, GMDB exhibits lower accuracy than VQAv2 ($0.6274$ vs.\ $0.7456$) despite similar MSR and slightly higher SDR. This indicates that although the model frequently attends to meaningful cross-modal cues, the inherent complexity of the rare disease domain constrains overall prediction accuracy. In image-text retrieval, MSCOCO achieves higher MSR ($0.5583$) while Flickr30K yields higher SDR ($0.5633$), reflecting dataset-specific characteristics: MSCOCO's literal captions encourage strong synergy on average, while Flickr30K's compositional captions require more frequent suppression of spurious alignments. These patterns confirm that MultiSHAP metrics meaningfully capture dataset-specific reasoning behaviours learned by multimodal models.

\subsection*{Interaction patterns correlate with phenotypic distinctiveness in rare diseases}
To further validate MultiSHAP's utility in medical applications, we examined three rare disease cohorts with differing levels of facial phenotypic distinctiveness: CdLS (highly distinctive facial features)~\cite{Deardorff2020CdLS}, Noonan syndrome (recognizable but variable characteristics)~\cite{Roberts2025Noonan}, and Angelman syndrome (primarily neurodevelopmental traits with less distinctive facial morphology)~\cite{Dagli2025Angelman}.

\begin{figure}[!htbp]
    \centering
    \includegraphics[width=0.9\linewidth]{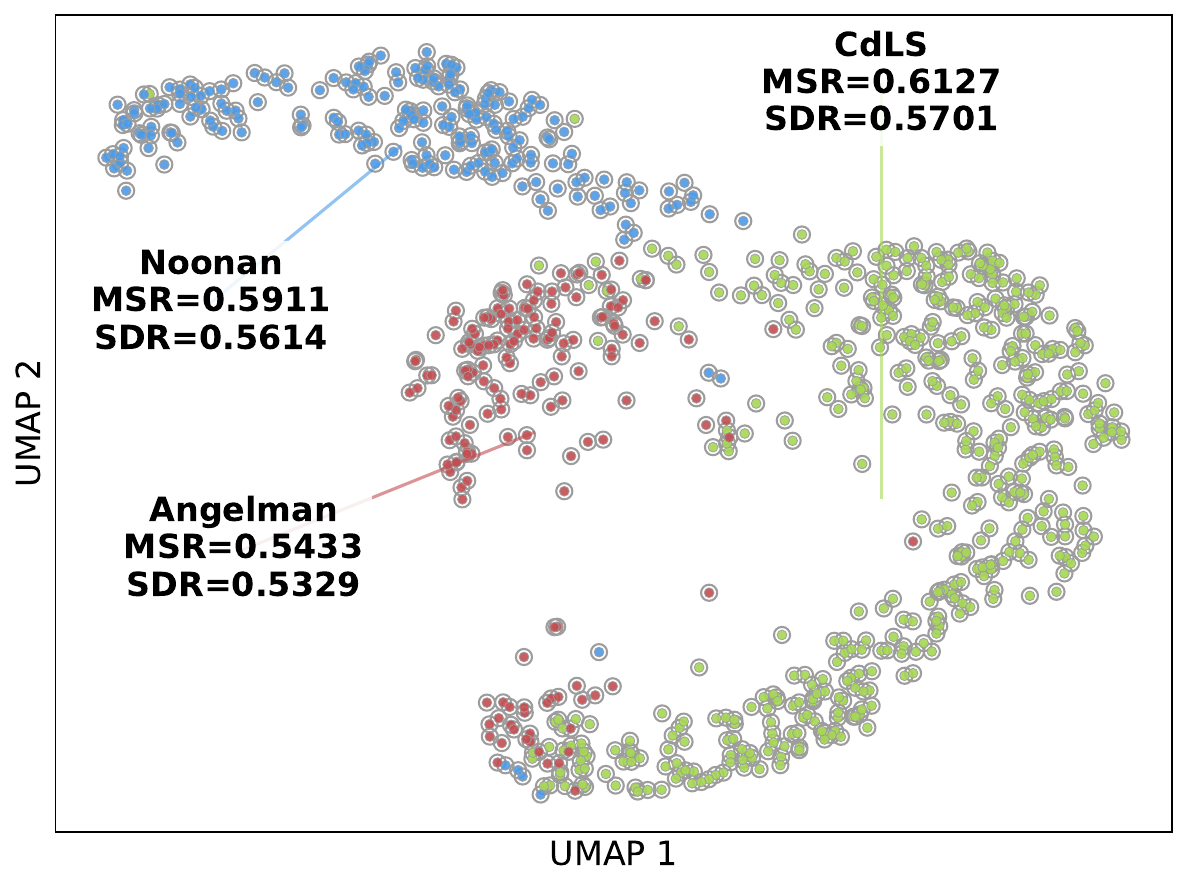}
    \caption{\textbf{MultiSHAP metrics correlate with phenotypic distinctiveness across rare disease cohorts.}
    UMAP visualization of patient image embeddings from three rare disease cohorts in the GestaltMatcher Database. Distinct clustering is observed for Cornelia de Lange syndrome (CdLS), Noonan syndrome, and Angelman syndrome. Dataset-level MultiSHAP statistics (inset) show that Mean Synergy Ratio (MSR) and Synergy Dominance Ratio (SDR) decrease with phenotypic distinctiveness: CdLS (most distinctive facial features) exhibits the strongest multimodal synergy (MSR $= 0.61$, SDR $= 0.57$), followed by Noonan syndrome (MSR $= 0.59$, SDR $= 0.56$), while Angelman syndrome (least distinctive facial morphology) shows the weakest synergy (MSR $= 0.54$, SDR $= 0.53$). This correlation suggests that cross-modal interactions contribute more strongly to predictions when facial phenotypes are more informative.}
    \label{fig:umap_embedding}
\end{figure}

UMAP visualization of patient image embeddings reveals distinct clustering for each cohort (Fig.~\ref{fig:umap_embedding}), suggesting that the model captures cohort-specific facial characteristics. Importantly, MultiSHAP metrics correlate with phenotypic distinctiveness: CdLS exhibits the strongest multimodal synergy (MSR $= 0.6127$, SDR $= 0.5701$), followed by Noonan syndrome (MSR $= 0.5911$, SDR $= 0.5614$), while Angelman syndrome shows the weakest synergy (MSR $= 0.5433$, SDR $= 0.5329$).

These results suggest that image-text interactions contribute more strongly to model predictions for cohorts with more distinctive facial phenotypes. This finding has important implications for clinical applications: MultiSHAP can help identify which diagnostic cases may benefit from stronger visual evidence integration and which may require additional clinical context beyond facial analysis.

\section*{Discussion}
\label{sec:discussion}

We have presented MultiSHAP, a unified framework for quantifying cross-modal interactions in multimodal AI models based on the Shapley Interaction Index from cooperative game theory. By computing fine-grained synergy and suppression scores between image patches and text tokens, MultiSHAP produces interpretable interaction matrices that reveal how visual and textual elements collaborate or conflict during model inference. Our analysis across visual question answering and image-text retrieval tasks identified four distinct interaction patterns: beneficial synergy supporting correct predictions, harmful synergy amplifying irrelevant evidence, helpful suppression filtering misleading cues, and detrimental suppression undermining accuracy. These patterns provide mechanistic insights into multimodal reasoning that go beyond what existing attribution methods can offer.

A key advantage of MultiSHAP over prior approaches is its ability to capture fine-grained patch-token interactions while remaining model-agnostic. Previous Shapley-based methods such as InterSHAP~\cite{Wenderoth_2025} and MM-SHAP~\cite{mmshap2023} quantify only modality-level contributions, treating entire images or text sequences as single features. While useful for understanding global modality importance, these approaches cannot reveal which specific visual regions interact with which textual concepts. Unimodal methods such as TokenSHAP~\cite{goldshmidt2024tokenshapinterpretinglargelanguage} and PixelSHAP~\cite{goldshmidt2025attentionpleasepixelshapreveals} provide fine-grained attributions within a single modality but do not capture cross-modal interactions. Attention-based visualizations~\cite{attentionmap} and gradient methods such as Grad-CAM~\cite{Selvaraju_2019} require access to model internals and often conflate correlation with causation. MultiSHAP addresses these limitations by providing counterfactual, axiomatic explanations at the patch-token level without requiring internal model access.

Our results demonstrate that interaction patterns carry diagnostic value beyond prediction accuracy. In rare disease diagnosis, we found that synergy-dominated interactions ($R_k > 0.5$) generally corresponded to cases where the model correctly integrated clinically relevant facial features with diagnostic queries, while suppression-dominated interactions often indicated failure modes where critical visual evidence was inappropriately down-weighted. Importantly, the correlation between MultiSHAP metrics and phenotypic distinctiveness across disease cohorts (Fig.~\ref{fig:umap_embedding}) suggests that these metrics capture meaningful properties of the underlying diagnostic task. Cohorts with more distinctive facial phenotypes (CdLS) exhibited stronger multimodal synergy than those with less distinctive features (Angelman syndrome), consistent with clinical expectations about the relative informativeness of facial analysis for different genetic conditions~\cite{Deardorff2020CdLS, Roberts2025Noonan, Dagli2025Angelman}.

The discovery that suppressive interactions can serve beneficial functions has implications for understanding multimodal reasoning. In the VQA example where the model correctly answered that not both dogs were white (Fig.~\ref{fig:main_examples}(c)), suppressive interactions with the white dog helped filter potentially misleading evidence. This suggests that well-functioning multimodal models may actively suppress irrelevant or contradictory information rather than simply amplifying relevant features. Conversely, the failure case where spurious synergy with colorful objects led to an incorrect colour prediction (Fig.~\ref{fig:main_examples}d) illustrates how positive interactions with visually salient but semantically irrelevant regions can override correct evidence. These findings highlight the importance of examining both synergistic and suppressive interactions when diagnosing model behaviour.

The dataset-level metrics we introduced: Mean Synergy Ratio (MSR) and Synergy Dominance Ratio (SDR), which provide complementary insights to standard performance metrics. Our observation that GMDB exhibits similar MSR and SDR to VQAv2 despite substantially lower accuracy suggests that the medical diagnosis model attends to meaningful cross-modal cues but faces inherent task complexity that limits predictive performance. Similarly, the different MSR-SDR profiles between MSCOCO (higher MSR) and Flickr30K (higher SDR) reflect dataset-specific caption styles: literal descriptions encourage consistent synergy, while compositional captions require more selective suppression of spurious alignments. These metrics could serve as diagnostic tools for model development, helping identify whether performance limitations stem from attention failures or task complexity.

Several limitations of the current work suggest directions for future research. First, the Monte Carlo estimation procedure requires $O(K \times m \times n)$ model evaluations per sample, which becomes computationally expensive for high-resolution images or long text sequences. Developing more efficient approximations, perhaps leveraging structured sampling or amortized inference, would improve scalability. Second, our current formulation considers only pairwise patch-token interactions; extending to higher-order interactions could reveal more complex reasoning patterns but would further increase computational cost. Third, while we demonstrated MultiSHAP on vision-language models, the framework could be extended to other modality combinations (audio-text, video-text) or to scenarios with more than two input modalities, though this would require careful consideration of how to define and visualize higher-dimensional interaction tensors.

Looking forward, MultiSHAP opens several avenues for improving multimodal AI systems. The ability to identify specific patch-token interactions that cause failures could inform targeted data augmentation or architectural modifications. Dataset-level metrics could guide model selection for deployment in specific domains. In clinical applications, MultiSHAP visualizations could support human-AI collaboration by highlighting which image regions and textual concepts the model considers most relevant, enabling clinicians to verify or override model reasoning. More broadly, the framework contributes to the growing toolkit for interpretable AI, addressing the critical need for transparency in multimodal systems deployed in high-stakes domains.
\backmatter

\bmhead{Supplementary information}




The online version of this article contains supplementary material, including:
(i) additional token-wise patch--token interaction heatmaps for all qualitative case studies;
(ii) supplementary tables reporting method comparisons, runtime analysis, and cohort-level interaction metrics; and
(iii) additional details on the experimental setup and evaluation.

\bmhead{Acknowledgements}
We thank Da Wu, Quan Nguyen, and Mian Umair Ahsan (Wang Genomics Lab, CHOP/Penn) for insightful comments and suggestions on the interpretation of multimodal AI models. We acknowledge the publicly available data resources used in this work, including the GestaltMatcher Database (GMDB), VQAv2, MS~COCO, and Flickr30K, and we thank the respective dataset creators, curators, and maintainers for making these resources available. This work was supported by NIH grant OD037960 and the CHOP Research Institute.


\section*{Declarations}

\subsection*{Funding}
This work was supported by NIH grant OD037960 and the CHOP Research Institute.

\subsection*{Competing interests}
The authors declare that they have no competing interests.

\subsection*{Ethics approval and consent to participate}
Not applicable. This study used publicly available benchmark datasets and controlled-access data obtained through the official data access process. No new human participant data were collected.

\subsection*{Consent for publication}
Not applicable.

\subsection*{Data availability}
Public datasets used in this study are available from their respective sources (VQAv2, MS~COCO, and Flickr30K). The GestaltMatcher Database (GMDB) is a controlled-access resource and is available upon approved application through the GMDB portal: \url{https://db.gestaltmatcher.org/}.

\subsection*{Materials availability}
Not applicable.

\subsection*{Code availability}
Code is available at \url{https://github.com/WGLab/MultiSHAP}.

\subsection*{Authors' contributions}
Z.W. and K.W. conceived the study. Z.W. implemented the method and performed experiments. Z.W. and K.W. analyzed results and wrote the manuscript. All authors reviewed and approved the final manuscript.










\newpage
\section*{Appendix}
\begin{appendices}

\section*{Token-wise Interaction Heatmaps}
\label{app:token_heatmaps}

In this section, we present full token-to-image patch interaction heatmaps for the qualitative examples. Each heatmap visualizes the Shapley interaction values $\Phi_{ij}$ between text token $t_j$ and image patch $p_i$. Warm colors (red) indicate synergistic interactions where token and patch mutually enhance the model's prediction, while cool colors (blue) represent suppressive interactions.

\subsection*{VQA Qualitative Analysis}

\subsubsection*{Example 3: Breakfast Recognition Success}
As shown in Supplementary figures, the token \textbf{"what"} creates broad synergistic interactions across food items, effectively priming visual search for objects on the plates. The token \textbf{"plates"} exhibits the strongest positive interactions with the physical plate regions, demonstrating accurate object grounding and spatial localization.

\subsubsection*{Example 5: Spurious Synergy and Reasoning Failure}
This failure case reveals the "right for the wrong reasons" phenomenon. As identified by MultiSHAP, the model's high confidence is driven by synergistic interactions between background context tokens and unrelated image patches, rather than the primary subject identified in the text~\cite{hou2025fairccafairrepresentation}.

\subsection*{Image-Text Retrieval Qualitative Analysis}

\subsubsection*{Example 8: Foil Detection and Mismatch Recognition}
In Figure~\ref{fig:token_patch_6}, the foil caption "A baby holding a watermelon in his left hand" reveals sophisticated mismatch recognition. The token \textbf{"watermelon"} shows strong suppressive interactions with the actual banana region, while spatial tokens correctly identify directional inconsistency.

\subsubsection*{Example 9: Category Grounding Success}
As shown in Figure~\ref{fig:token_patch_7}, the caption "These are some very large onions" demonstrates precise category grounding. The token \textbf{"onions"} creates strong positive interactions specifically in the onion regions, while modifiers like \textbf{"large"} provide appropriate semantic support.

\subsubsection*{Example 10: Category Mismatch Recognition}
In Figure~\ref{fig:token_patch_8}, the foil caption "These are some very large watermelons" exhibits the model's ability to reject incorrect labels. The token \textbf{"watermelons"} produces strong suppressive interactions in the spatial regions previously associated with "onions".

\clearpage
\section*{Supplementary Materials}
\label{app:extended_data}

\renewcommand{\thetable}{S\arabic{table}}
\setcounter{table}{0}

\renewcommand{\thefigure}{S\arabic{figure}}

\begin{table}[ht]
\centering
\small
\setlength{\tabcolsep}{6pt}
\begin{tabular}{lcccll}
\toprule
\textbf{Method} & \textbf{Multimodal} & \textbf{Agnostic} & \textbf{Granularity} & \textbf{Mechanism} & \textbf{Extra Req.} \\
\midrule
TokenSHAP~\cite{goldshmidt2024tokenshapinterpretinglargelanguage} & \xmark & \cmark & Token & Shapley Value & None \\
PixelSHAP~\cite{goldshmidt2025attentionpleasepixelshapreveals} & \xmark & \cmark & Pixel/Region & Shapley Value & Seg. Model \\
InterSHAP~\cite{Wenderoth_2025} & \cmark & \cmark & Modality & SII & None \\
Grad-CAM~\cite{selvaraju2017grad} & \cmark & \xmark & Region & Gradients & Model Access \\
\textbf{MultiSHAP} & \cmark & \cmark & Patch$\times$Token & SII & \textbf{None} \\
\bottomrule
\end{tabular}
\vspace{2mm}
\raggedright
\small
\caption{\textbf{Comparison of MultiSHAP with state-of-the-art attribution methods.} Unlike prior work focusing on unimodal importance or coarse modality-level interactions, MultiSHAP quantifies interactions at the atomic level (patches and tokens) without architectural dependencies. SII, Shapley Interaction Index.}
\label{tab:comparison_extended}
\end{table}

\begin{table}[ht]
\centering
\small
\setlength{\tabcolsep}{8pt}
\begin{tabular}{cccc}
\toprule
$\mathbf{K}$ & \textbf{Mean $\pm$ Std (s)} & \textbf{Total Time (500 samples)} & \textbf{Relative Slowdown} \\
\midrule
32  & $17.5 \pm 0.8$  & 2.43 h & $1.0\times$ \\
64  & $37.2 \pm 1.3$  & 5.17 h & $2.1\times$ \\
128 & $70.0 \pm 2.9$  & 9.72 h & $4.0\times$ \\
\bottomrule
\end{tabular}
\vspace{2mm}
\raggedright
\small
\caption{\textbf{MultiSHAP runtime analysis on hardware.} Benchmarked on an Apple M2 Max (32 GB RAM) using the ViLT-B/32 architecture on VQAv2. Each entry averages three independent runs. $K$ denotes the sampling budget for Monte Carlo estimation.}
\label{tab:runtime}
\end{table}

\begin{table}[htbp!]
\centering
\small
\setlength{\tabcolsep}{10pt}
\begin{tabular}{lccc}
\toprule
\textbf{Disease Cohort (GMDB)} & \textbf{MSR} & \textbf{SDR} & \textbf{Sample Size ($N$)} \\
\midrule
CdLS (Distinctive)           & $0.6127$ & $0.5701$ & $447$ \\
Noonan (Variable)            & $0.5911$ & $0.5614$ & $207$ \\
Angelman (Non-specific)      & $0.5433$ & $0.5329$ & $160$  \\
\bottomrule
\end{tabular}
\vspace{2mm}
\raggedright
\small
\caption{\textbf{MultiSHAP interaction metrics across rare disease cohorts.} Comparison of Mean Synergy Ratio (MSR) and Synergy Dominance Ratio (SDR) across syndromes in the GestaltMatcher Database (GMDB) with varying phenotypic distinctiveness.}
\label{tab:cohort}
\end{table}

\begin{figure*}[t]
\centering
\includegraphics[width=0.95\textwidth]{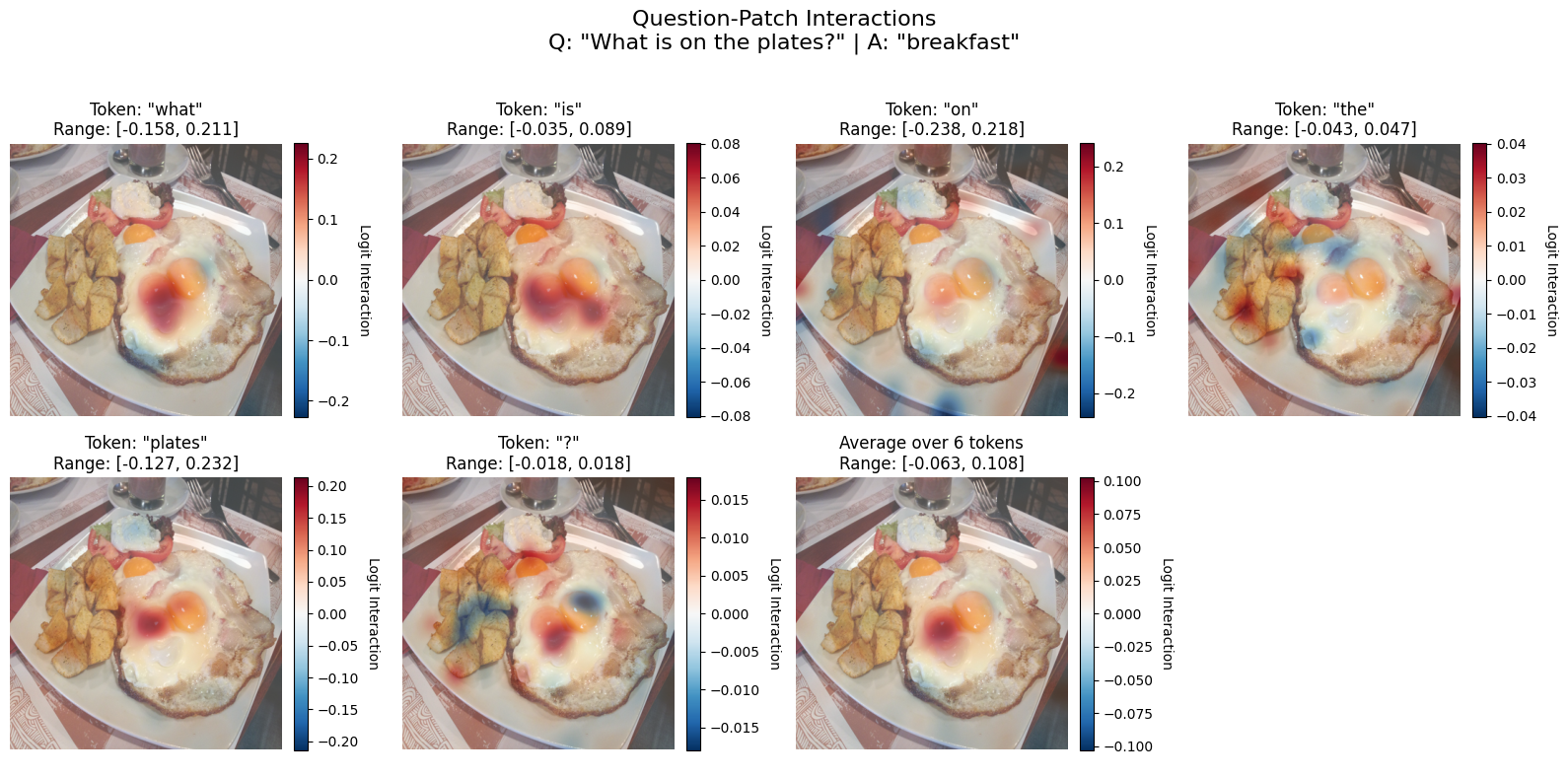}
\caption{\textbf{Token-level interaction heatmaps for VQA Example 3.} \textbf{Question}: "What is on the plates?" \textbf{Answer}: "breakfast" (correct). This successful case demonstrates ideal synergistic patterns where content words create strong positive interactions with semantically relevant food regions, while spatial tokens properly bind objects to locations.}
\label{fig:token_patch_1}
\end{figure*}

\begin{figure*}[t]
\centering
\includegraphics[width=0.95\textwidth]{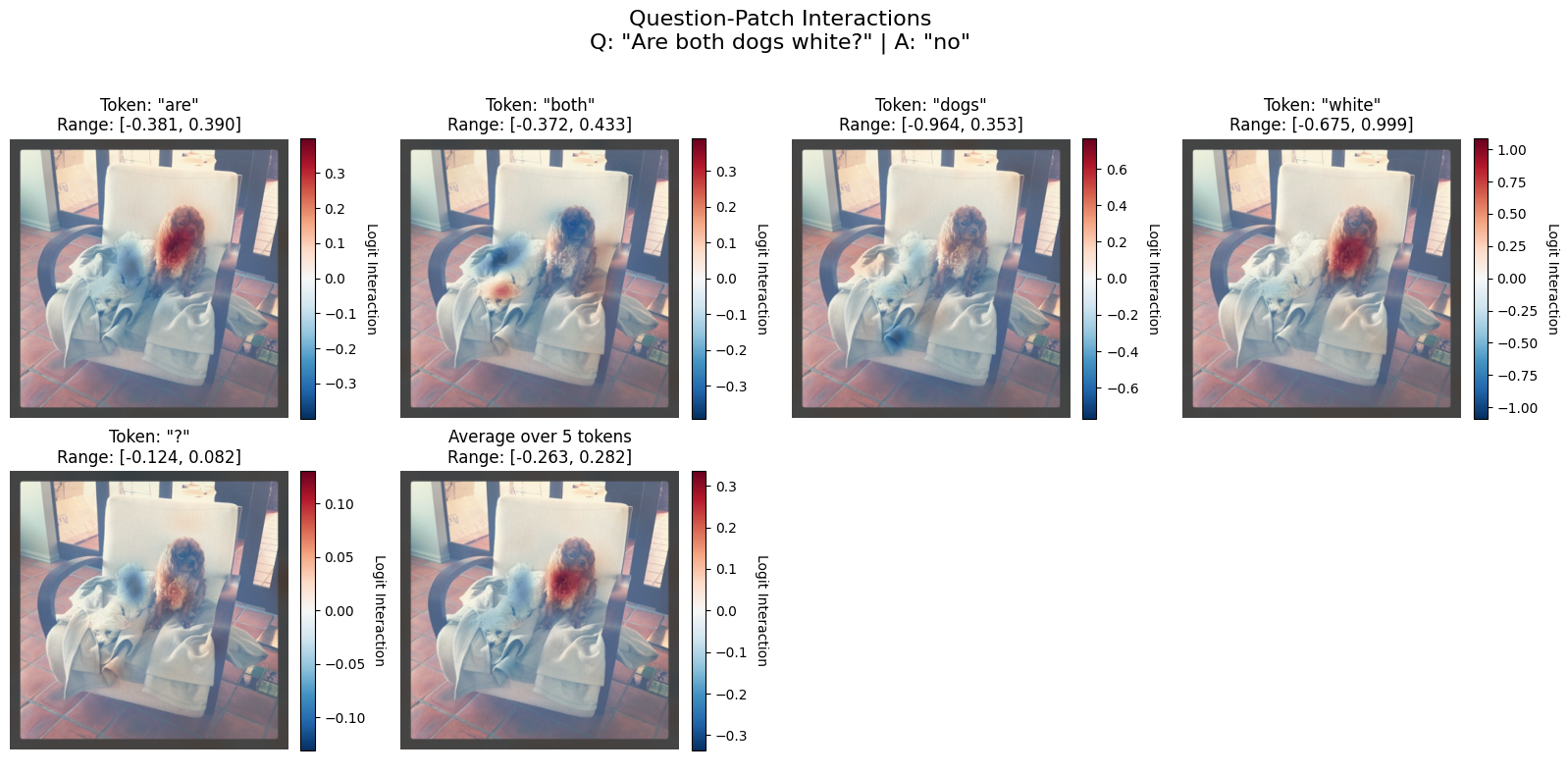}
\caption{\textbf{Token-level interaction heatmaps for VQA Example 4. }\textbf{Question}: "Are both dogs white?" \textbf{Answer}: "no" (correct). This case demonstrates how suppressive interactions can strategically filter misleading evidence, with the token "white" showing negative interactions with the white dog region to support the correct negative answer.}
\label{fig:token_patch_2}
\end{figure*}

\begin{figure*}[t]
\centering
\includegraphics[width=0.95\textwidth]{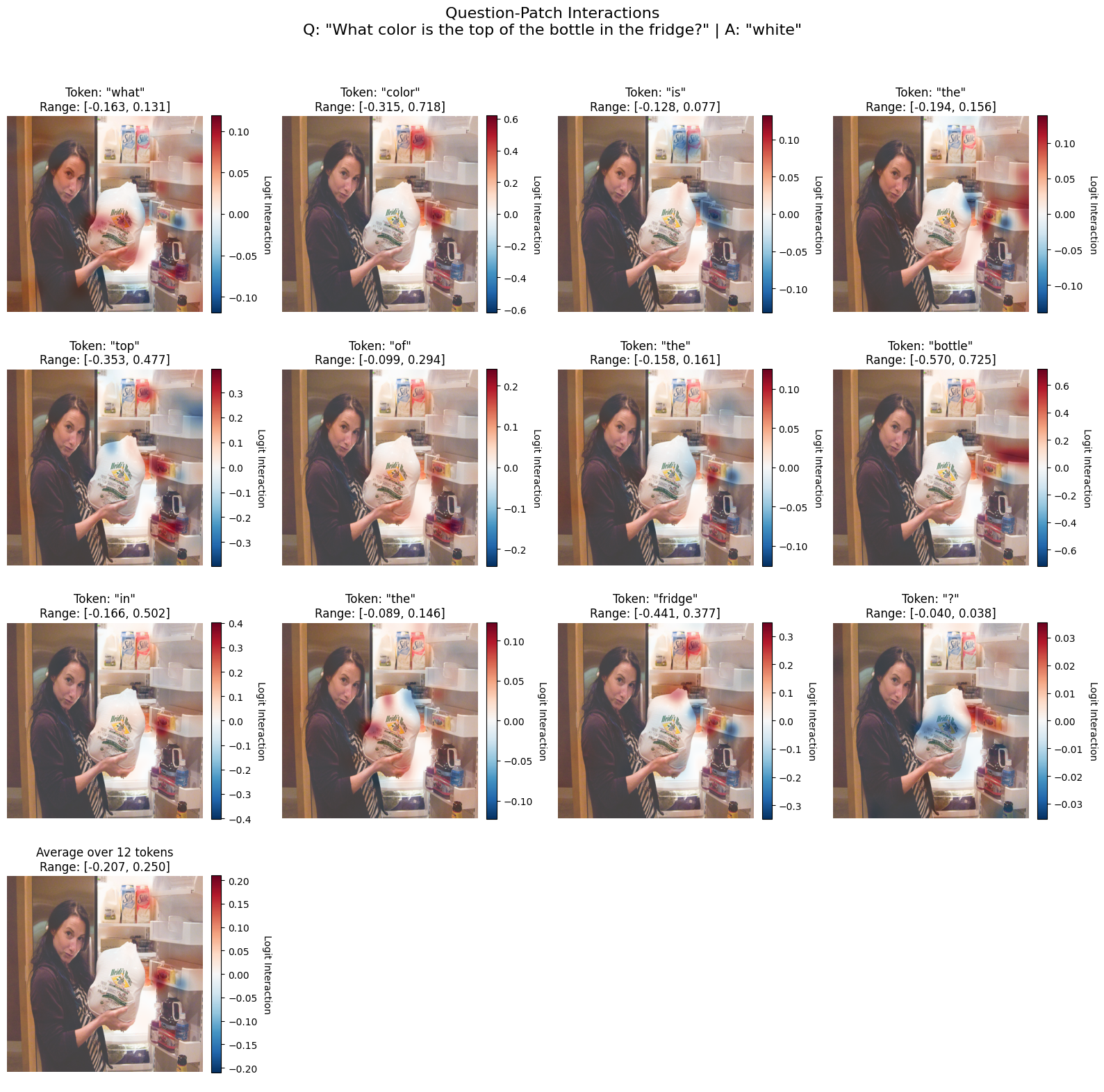}
\caption{\textbf{Token-level interaction heatmaps for VQA Example 5.} \textbf{Question}: "What color is the top of the bottle in the fridge?" \textbf{Answer}: "white" (incorrect, should be white). This failure case reveals how spurious positive interactions with visually salient but semantically irrelevant colorful objects can mislead the model away from the correct white bottle cap.}
\label{fig:token_patch_3}
\end{figure*}

\begin{figure*}[t]
\centering
\includegraphics[width=0.95\textwidth]{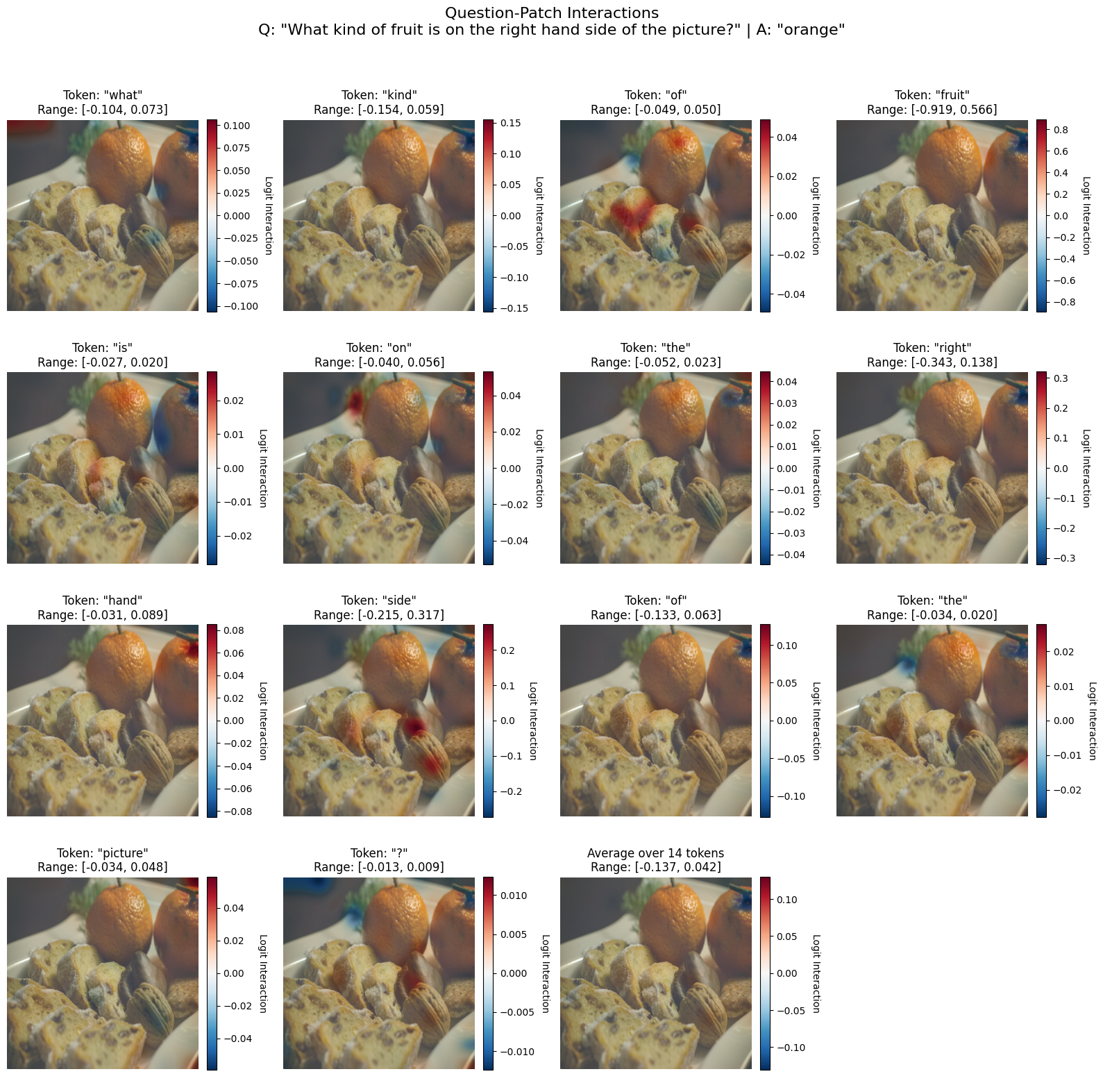}
\caption{\textbf{Token-level interaction heatmaps for VQA Example 6. }\textbf{Question}: "What kind of fruit is on the right hand side of the picture?" \textbf{Answer}: "orange" (incorrect, should be orange). This case shows how suppressive interactions with correct spatial regions can undermine accurate reasoning, with spatial tokens showing negative rather than positive interactions with the target orange fruit.}
\label{fig:token_patch_4}
\end{figure*}

\begin{figure*}[t]
\centering
\includegraphics[width=0.95\textwidth]{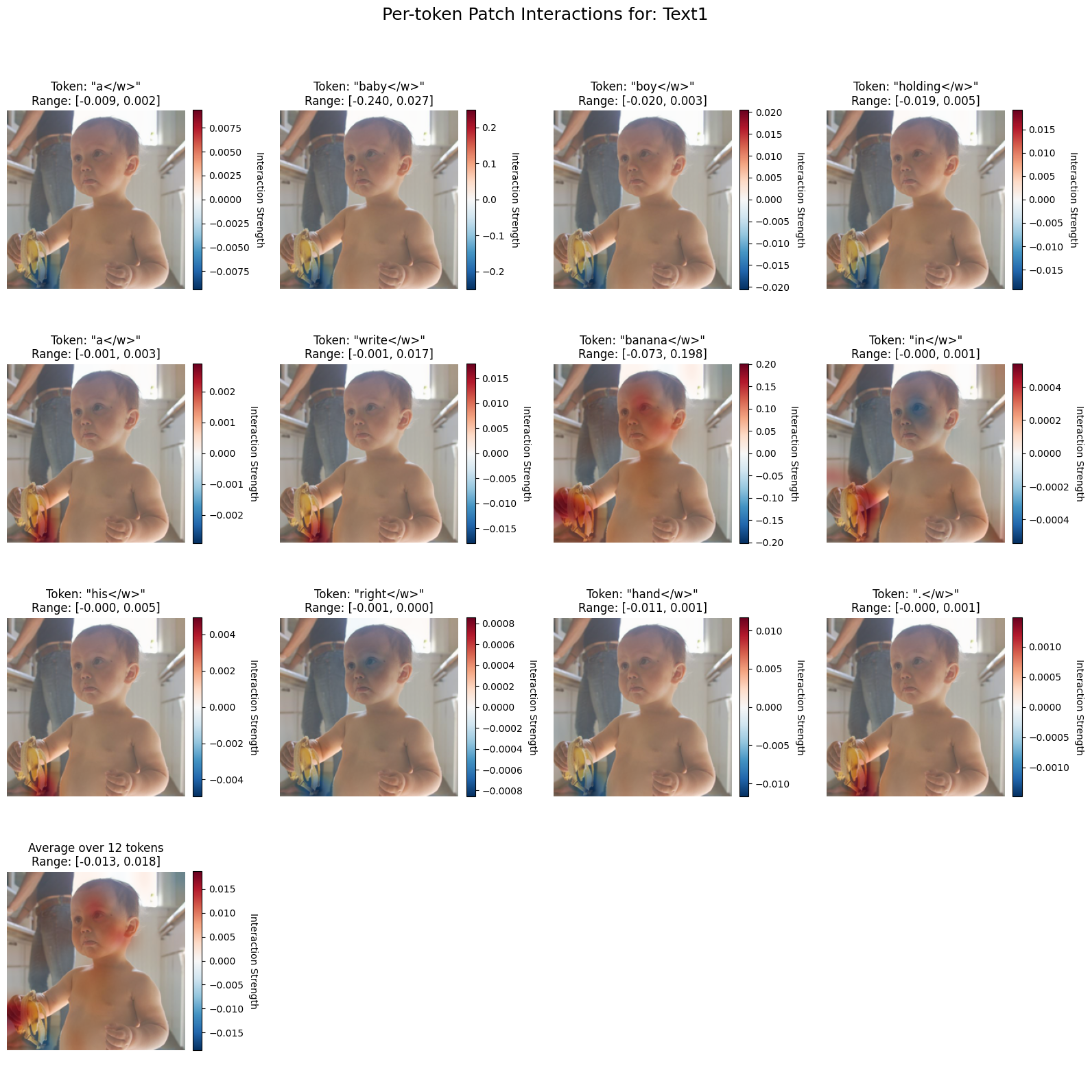}
\caption{\textbf{Token-level interaction heatmaps for Image-Text Retrieval Example 7.} \textbf{Caption}: "A baby holding a banana in his right hand" (ground truth). This successful case shows precise object-spatial grounding with "banana" creating strong positive interactions in the correct hand region and spatial tokens accurately localizing to the right side of the image.}
\label{fig:token_patch_5}
\end{figure*}

\begin{figure*}[t]
\centering
\includegraphics[width=0.95\textwidth]{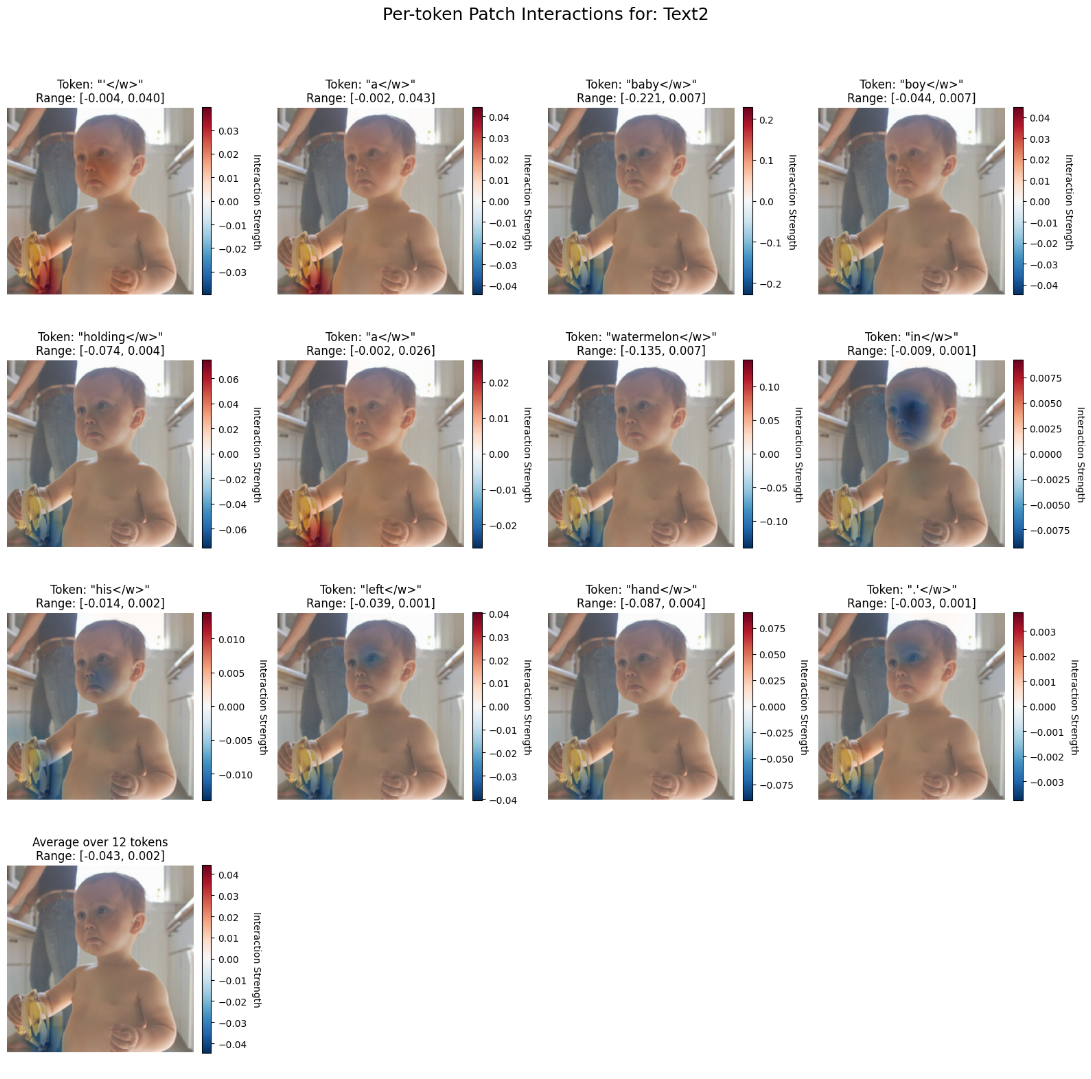}
\caption{\textbf{Token-level interaction heatmaps for Image-Text Retrieval Example 8.} \textbf{Caption}: "A baby holding a watermelon in his left hand" (foil). This foil detection case reveals sophisticated mismatch recognition with "watermelon" showing strong suppressive interactions with the actual banana region and spatial tokens correctly identifying directional inconsistency.}
\label{fig:token_patch_6}
\end{figure*}

\begin{figure*}[t]
\centering
\includegraphics[width=0.95\textwidth]{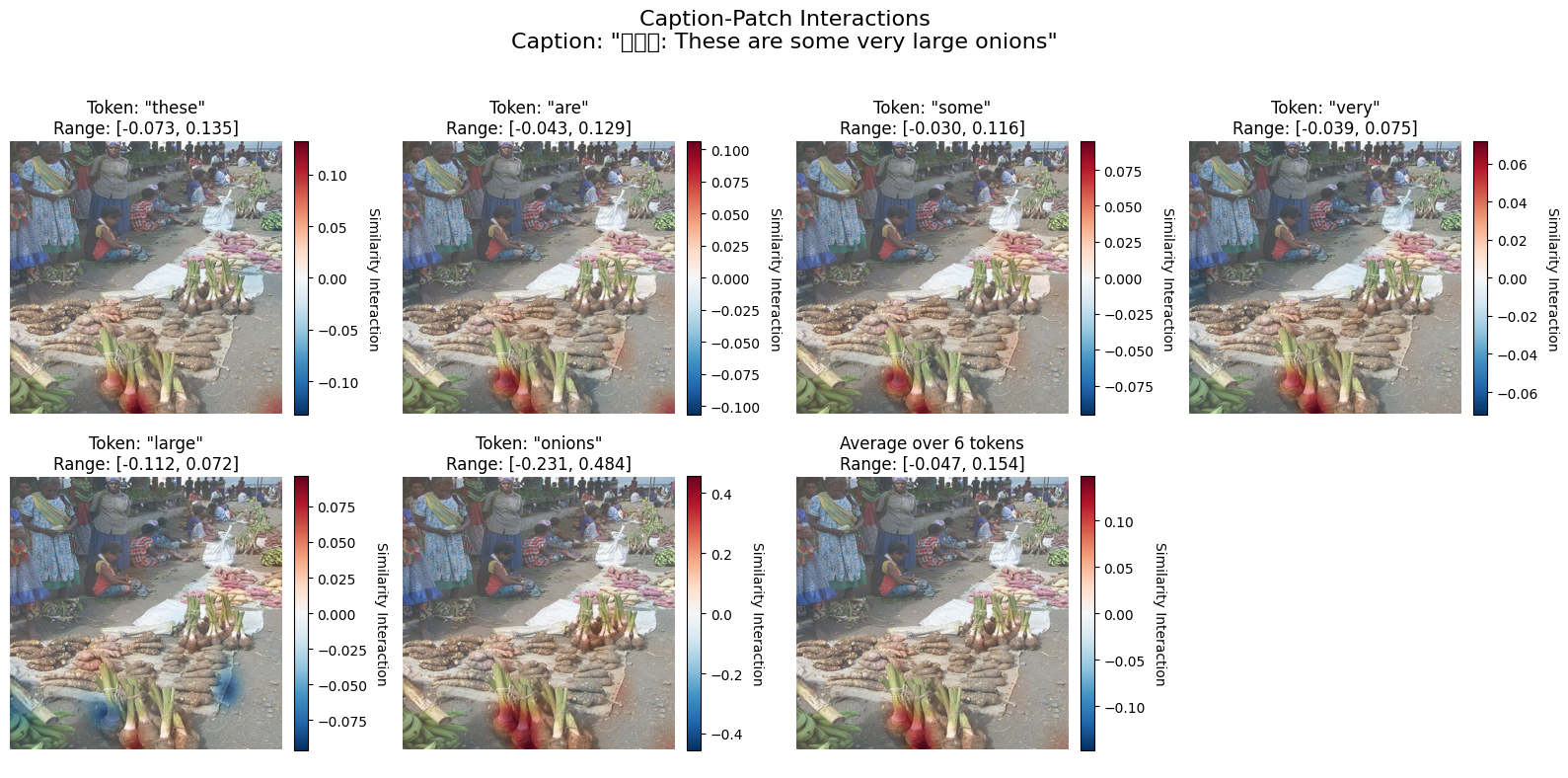}
\caption{\textbf{Token-level interaction heatmaps for Image-Text Retrieval Example 9.} \textbf{Caption}: "These are some very large onions" (ground truth). This case demonstrates precise category grounding with "onions" creating strong positive interactions specifically in the onion regions while modifier tokens like "large" and "very" provide appropriate semantic support.}
\label{fig:token_patch_7}
\end{figure*}

\begin{figure*}[t]
\centering
\includegraphics[width=0.95\textwidth]{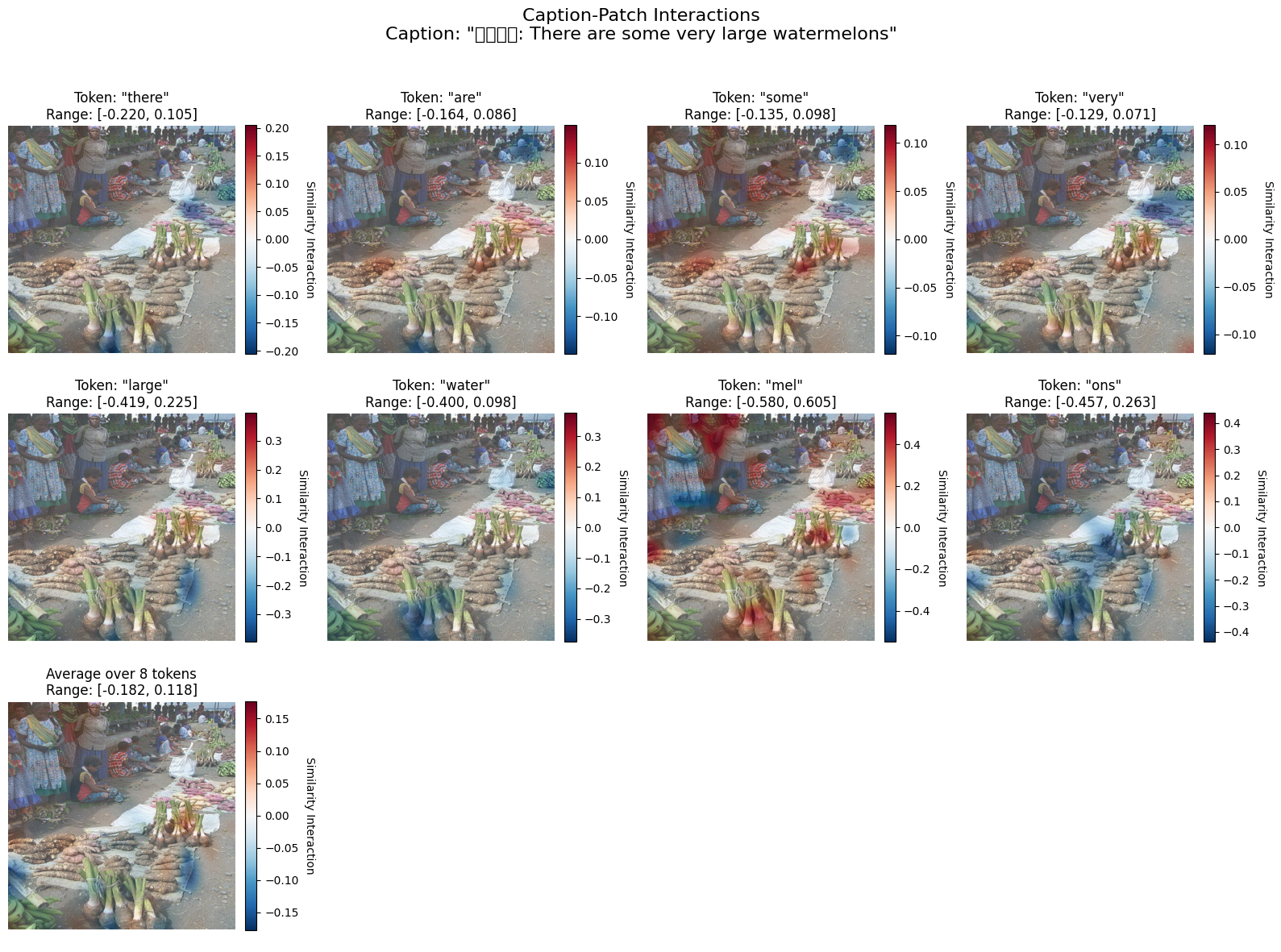}
\caption{\textbf{Token-level interaction heatmaps for Image-Text Retrieval Example 10.} \textbf{Caption}: "These are some very large watermelons" (foil). This category mismatch case shows the model's ability to reject incorrect category labels with "watermelons" creating strong suppressive interactions in the same spatial regions that previously showed positive interactions for "onions".}
\label{fig:token_patch_8}
\end{figure*}
\end{appendices}

\clearpage
\bibliography{sn-bibliography}


\begin{thebibliography}{31}
\ifx \bisbn   \undefined \def \bisbn  #1{ISBN #1}\fi
\ifx \binits  \undefined \def \binits#1{#1}\fi
\ifx \bauthor  \undefined \def \bauthor#1{#1}\fi
\ifx \batitle  \undefined \def \batitle#1{#1}\fi
\ifx \bjtitle  \undefined \def \bjtitle#1{#1}\fi
\ifx \bvolume  \undefined \def \bvolume#1{\textbf{#1}}\fi
\ifx \byear  \undefined \def \byear#1{#1}\fi
\ifx \bissue  \undefined \def \bissue#1{#1}\fi
\ifx \bfpage  \undefined \def \bfpage#1{#1}\fi
\ifx \blpage  \undefined \def \blpage #1{#1}\fi
\ifx \burl  \undefined \def \burl#1{\textsf{#1}}\fi
\ifx \doiurl  \undefined \def \doiurl#1{\url{https://doi.org/#1}}\fi
\ifx \betal  \undefined \def \betal{\textit{et al.}}\fi
\ifx \binstitute  \undefined \def \binstitute#1{#1}\fi
\ifx \binstitutionaled  \undefined \def \binstitutionaled#1{#1}\fi
\ifx \bctitle  \undefined \def \bctitle#1{#1}\fi
\ifx \beditor  \undefined \def \beditor#1{#1}\fi
\ifx \bpublisher  \undefined \def \bpublisher#1{#1}\fi
\ifx \bbtitle  \undefined \def \bbtitle#1{#1}\fi
\ifx \bedition  \undefined \def \bedition#1{#1}\fi
\ifx \bseriesno  \undefined \def \bseriesno#1{#1}\fi
\ifx \blocation  \undefined \def \blocation#1{#1}\fi
\ifx \bsertitle  \undefined \def \bsertitle#1{#1}\fi
\ifx \bsnm \undefined \def \bsnm#1{#1}\fi
\ifx \bsuffix \undefined \def \bsuffix#1{#1}\fi
\ifx \bparticle \undefined \def \bparticle#1{#1}\fi
\ifx \barticle \undefined \def \barticle#1{#1}\fi
\bibcommenthead
\ifx \bconfdate \undefined \def \bconfdate #1{#1}\fi
\ifx \botherref \undefined \def \botherref #1{#1}\fi
\ifx \url \undefined \def \url#1{\textsf{#1}}\fi
\ifx \bchapter \undefined \def \bchapter#1{#1}\fi
\ifx \bbook \undefined \def \bbook#1{#1}\fi
\ifx \bcomment \undefined \def \bcomment#1{#1}\fi
\ifx \oauthor \undefined \def \oauthor#1{#1}\fi
\ifx \citeauthoryear \undefined \def \citeauthoryear#1{#1}\fi
\ifx \endbibitem  \undefined \def \endbibitem {}\fi
\ifx \bconflocation  \undefined \def \bconflocation#1{#1}\fi
\ifx \arxivurl  \undefined \def \arxivurl#1{\textsf{#1}}\fi
\csname PreBibitemsHook\endcsname

\bibitem[\protect\citeauthoryear{Antol et~al.}{2015}]{antol2015vqa}
\begin{bchapter}
\bauthor{\bsnm{Antol}, \binits{S.}},
\bauthor{\bsnm{Agrawal}, \binits{A.}},
\bauthor{\bsnm{Lu}, \binits{J.}},
\bauthor{\bsnm{Mitchell}, \binits{M.}},
\bauthor{\bsnm{Batra}, \binits{D.}},
\bauthor{\bsnm{Lawrence~Zitnick}, \binits{C.}},
\bauthor{\bsnm{Parikh}, \binits{D.}}:
\bctitle{Vqa: Visual question answering}.
In: \bbtitle{Proceedings of the IEEE International Conference on Computer Vision},
pp. \bfpage{2425}--\blpage{2433}
(\byear{2015})
\end{bchapter}
\endbibitem

\bibitem[\protect\citeauthoryear{Goyal et~al.}{2017}]{goyal2017making}
\begin{bchapter}
\bauthor{\bsnm{Goyal}, \binits{Y.}},
\bauthor{\bsnm{Khot}, \binits{T.}},
\bauthor{\bsnm{Summers-Stay}, \binits{D.}},
\bauthor{\bsnm{Batra}, \binits{D.}},
\bauthor{\bsnm{Parikh}, \binits{D.}}:
\bctitle{Making the v in vqa matter: Elevating the role of image understanding in visual question answering}.
In: \bbtitle{Proceedings of the IEEE Conference on Computer Vision and Pattern Recognition},
pp. \bfpage{6904}--\blpage{6913}
(\byear{2017})
\end{bchapter}
\endbibitem

\bibitem[\protect\citeauthoryear{Lin et~al.}{2014}]{lin2014microsoft}
\begin{bchapter}
\bauthor{\bsnm{Lin}, \binits{T.-Y.}},
\bauthor{\bsnm{Maire}, \binits{M.}},
\bauthor{\bsnm{Belongie}, \binits{S.}},
\bauthor{\bsnm{Hays}, \binits{J.}},
\bauthor{\bsnm{Perona}, \binits{P.}},
\bauthor{\bsnm{Ramanan}, \binits{D.}},
\bauthor{\bsnm{Doll{\'a}r}, \binits{P.}},
\bauthor{\bsnm{Zitnick}, \binits{C.L.}}:
\bctitle{Microsoft coco: Common objects in context}.
In: \bbtitle{European Conference on Computer Vision},
pp. \bfpage{740}--\blpage{755}
(\byear{2014})
\end{bchapter}
\endbibitem

\bibitem[\protect\citeauthoryear{Young et~al.}{2014}]{young2014image}
\begin{barticle}
\bauthor{\bsnm{Young}, \binits{P.}},
\bauthor{\bsnm{Lai}, \binits{A.}},
\bauthor{\bsnm{Hodosh}, \binits{M.}},
\bauthor{\bsnm{Hockenmaier}, \binits{J.}}:
\batitle{From image descriptions to visual denotations: New similarity metrics for semantic inference over event descriptions}.
\bjtitle{Transactions of the Association for Computational Linguistics}
\bvolume{2},
\bfpage{67}--\blpage{78}
(\byear{2014})
\end{barticle}
\endbibitem

\bibitem[\protect\citeauthoryear{Radford et~al.}{2021}]{radford2021learningtransferablevisualmodels}
\begin{botherref}
\oauthor{\bsnm{Radford}, \binits{A.}},
\oauthor{\bsnm{Kim}, \binits{J.W.}},
\oauthor{\bsnm{Hallacy}, \binits{C.}},
\oauthor{\bsnm{Ramesh}, \binits{A.}},
\oauthor{\bsnm{Goh}, \binits{G.}},
\oauthor{\bsnm{Agarwal}, \binits{S.}},
\oauthor{\bsnm{Sastry}, \binits{G.}},
\oauthor{\bsnm{Askell}, \binits{A.}},
\oauthor{\bsnm{Mishkin}, \binits{P.}},
\oauthor{\bsnm{Clark}, \binits{J.}},
\oauthor{\bsnm{Krueger}, \binits{G.}},
\oauthor{\bsnm{Sutskever}, \binits{I.}}:
Learning Transferable Visual Models From Natural Language Supervision
(2021).
\url{https://arxiv.org/abs/2103.00020}
\end{botherref}
\endbibitem

\bibitem[\protect\citeauthoryear{Kim et~al.}{2021}]{kim2021viltvisionandlanguagetransformerconvolution}
\begin{botherref}
\oauthor{\bsnm{Kim}, \binits{W.}},
\oauthor{\bsnm{Son}, \binits{B.}},
\oauthor{\bsnm{Kim}, \binits{I.}}:
ViLT: Vision-and-Language Transformer Without Convolution or Region Supervision
(2021).
\url{https://arxiv.org/abs/2102.03334}
\end{botherref}
\endbibitem

\bibitem[\protect\citeauthoryear{Liu et~al.}{2023}]{liu2023improvedllava}
\begin{botherref}
\oauthor{\bsnm{Liu}, \binits{H.}},
\oauthor{\bsnm{Li}, \binits{C.}},
\oauthor{\bsnm{Li}, \binits{Y.}},
\oauthor{\bsnm{Lee}, \binits{Y.J.}}:
Improved Baselines with Visual Instruction Tuning.
arXiv:2310.03744
(2023)
\end{botherref}
\endbibitem

\bibitem[\protect\citeauthoryear{Rodis et~al.}{2024}]{rodis2024multimodalexplainableartificialintelligence}
\begin{botherref}
\oauthor{\bsnm{Rodis}, \binits{N.}},
\oauthor{\bsnm{Sardianos}, \binits{C.}},
\oauthor{\bsnm{Radoglou-Grammatikis}, \binits{P.}},
\oauthor{\bsnm{Sarigiannidis}, \binits{P.}},
\oauthor{\bsnm{Varlamis}, \binits{I.}},
\oauthor{\bsnm{Papadopoulos}, \binits{G.T.}}:
Multimodal Explainable Artificial Intelligence: A Comprehensive Review of Methodological Advances and Future Research Directions
(2024).
\url{https://arxiv.org/abs/2306.05731}
\end{botherref}
\endbibitem

\bibitem[\protect\citeauthoryear{Xiao et~al.}{2025}]{xiao2025restoringcalibrationalignedlarge}
\begin{botherref}
\oauthor{\bsnm{Xiao}, \binits{J.}},
\oauthor{\bsnm{Hou}, \binits{B.}},
\oauthor{\bsnm{Wang}, \binits{Z.}},
\oauthor{\bsnm{Jin}, \binits{R.}},
\oauthor{\bsnm{Long}, \binits{Q.}},
\oauthor{\bsnm{Su}, \binits{W.J.}},
\oauthor{\bsnm{Shen}, \binits{L.}}:
Restoring Calibration for Aligned Large Language Models: A Calibration-Aware Fine-Tuning Approach
(2025).
\url{https://arxiv.org/abs/2505.01997}
\end{botherref}
\endbibitem

\bibitem[\protect\citeauthoryear{Huang et~al.}{2022}]{huang2022interpretability}
\begin{botherref}
\oauthor{\bsnm{Huang}, \binits{Z.}},
\oauthor{\bsnm{Li}, \binits{F.}},
\oauthor{\bsnm{Wang}, \binits{Z.}},
\oauthor{\bsnm{Wang}, \binits{Z.}}:
Interpretability of deep learning.
Int. J. Future Comput. Commun
\textbf{11}(10.18178)
(2022)
\end{botherref}
\endbibitem

\bibitem[\protect\citeauthoryear{Wu et~al.}{2025a}]{wu2025mint}
\begin{bchapter}
\bauthor{\bsnm{Wu}, \binits{D.}},
\bauthor{\bsnm{Wang}, \binits{Z.}},
\bauthor{\bsnm{Nguyen}, \binits{Q.M.}},
\bauthor{\bsnm{Xu}, \binits{Z.}},
\bauthor{\bsnm{Wang}, \binits{K.}}:
\bctitle{{MINT}: Multimodal integrated knowledge transfer to large language models through preference optimization with biomedical applications}.
In: \bbtitle{ICML 2025 Generative AI and Biology (GenBio) Workshop}
(\byear{2025}).
\burl{https://openreview.net/forum?id=yhvHryyw80}
\end{bchapter}
\endbibitem

\bibitem[\protect\citeauthoryear{Wu et~al.}{2025b}]{wu2025integratingchainofthoughtretrievalaugmented}
\begin{botherref}
\oauthor{\bsnm{Wu}, \binits{D.}},
\oauthor{\bsnm{Wang}, \binits{Z.}},
\oauthor{\bsnm{Nguyen}, \binits{Q.}},
\oauthor{\bsnm{Wang}, \binits{K.}}:
Integrating Chain-of-Thought and Retrieval Augmented Generation Enhances Rare Disease Diagnosis from Clinical Notes
(2025).
\url{https://arxiv.org/abs/2503.12286}
\end{botherref}
\endbibitem

\bibitem[\protect\citeauthoryear{Nguyen et~al.}{2025}]{10.1145/3765612.3767763}
\begin{bbook}
\bauthor{\bsnm{Nguyen}, \binits{Q.M.}},
\bauthor{\bsnm{Ahsan}, \binits{M.U.}},
\bauthor{\bsnm{Wang}, \binits{Z.}},
\bauthor{\bsnm{Wang}, \binits{K.}}:
\bbtitle{PhenoGPT2: A Multimodal Fine-tuned Large Language Models for Phenotype Extraction and Normalization from Clinical Text and Facial Images}.
\bpublisher{Association for Computing Machinery},
\blocation{New York, NY, USA}
(\byear{2025}).
\burl{https://doi.org/10.1145/3765612.3767763}
\end{bbook}
\endbibitem

\bibitem[\protect\citeauthoryear{Hou et~al.}{2025}]{hou2025fairccafairrepresentation}
\begin{botherref}
\oauthor{\bsnm{Hou}, \binits{B.}},
\oauthor{\bsnm{Wang}, \binits{Z.}},
\oauthor{\bsnm{Zhou}, \binits{Z.}},
\oauthor{\bsnm{Tong}, \binits{B.}},
\oauthor{\bsnm{Wang}, \binits{Z.}},
\oauthor{\bsnm{Bao}, \binits{J.}},
\oauthor{\bsnm{Duong-Tran}, \binits{D.}},
\oauthor{\bsnm{Long}, \binits{Q.}},
\oauthor{\bsnm{Shen}, \binits{L.}}:
Fair CCA for Fair Representation Learning: An ADNI Study
(2025).
\url{https://arxiv.org/abs/2507.09382}
\end{botherref}
\endbibitem

\bibitem[\protect\citeauthoryear{Selvaraju et~al.}{2019}]{Selvaraju_2019}
\begin{barticle}
\bauthor{\bsnm{Selvaraju}, \binits{R.R.}},
\bauthor{\bsnm{Cogswell}, \binits{M.}},
\bauthor{\bsnm{Das}, \binits{A.}},
\bauthor{\bsnm{Vedantam}, \binits{R.}},
\bauthor{\bsnm{Parikh}, \binits{D.}},
\bauthor{\bsnm{Batra}, \binits{D.}}:
\batitle{Grad-cam: Visual explanations from deep networks via gradient-based localization}.
\bjtitle{International Journal of Computer Vision}
\bvolume{128}(\bissue{2}),
\bfpage{336}--\blpage{359}
(\byear{2019})
\doiurl{10.1007/s11263-019-01228-7}
\end{barticle}
\endbibitem

\bibitem[\protect\citeauthoryear{Chefer et~al.}{2021}]{attentionmap}
\begin{botherref}
\oauthor{\bsnm{Chefer}, \binits{H.}},
\oauthor{\bsnm{Gur}, \binits{S.}},
\oauthor{\bsnm{Wolf}, \binits{L.}}:
Transformer Interpretability Beyond Attention Visualization
(2021).
\url{https://arxiv.org/abs/2012.09838}
\end{botherref}
\endbibitem

\bibitem[\protect\citeauthoryear{Wenderoth et~al.}{2025}]{Wenderoth_2025}
\begin{barticle}
\bauthor{\bsnm{Wenderoth}, \binits{L.}},
\bauthor{\bsnm{Hemker}, \binits{K.}},
\bauthor{\bsnm{Simidjievski}, \binits{N.}},
\bauthor{\bsnm{Jamnik}, \binits{M.}}:
\batitle{Measuring cross-modal interactions in multimodal models}.
\bjtitle{Proceedings of the AAAI Conference on Artificial Intelligence}
\bvolume{39}(\bissue{20}),
\bfpage{21501}--\blpage{21509}
(\byear{2025})
\doiurl{10.1609/aaai.v39i20.35452}
\end{barticle}
\endbibitem

\bibitem[\protect\citeauthoryear{Goldshmidt and Horovicz}{2024}]{goldshmidt2024tokenshapinterpretinglargelanguage}
\begin{botherref}
\oauthor{\bsnm{Goldshmidt}, \binits{R.}},
\oauthor{\bsnm{Horovicz}, \binits{M.}}:
TokenSHAP: Interpreting Large Language Models with Monte Carlo Shapley Value Estimation
(2024).
\url{https://arxiv.org/abs/2407.10114}
\end{botherref}
\endbibitem

\bibitem[\protect\citeauthoryear{Goldshmidt}{2025}]{goldshmidt2025attentionpleasepixelshapreveals}
\begin{botherref}
\oauthor{\bsnm{Goldshmidt}, \binits{R.}}:
Attention, Please! PixelSHAP Reveals What Vision-Language Models Actually Focus On
(2025).
\url{https://arxiv.org/abs/2503.06670}
\end{botherref}
\endbibitem

\bibitem[\protect\citeauthoryear{Parcalabescu and Frank}{2023}]{mmshap2023}
\begin{bchapter}
\bauthor{\bsnm{Parcalabescu}, \binits{L.}},
\bauthor{\bsnm{Frank}, \binits{A.}}:
\bctitle{Mm-shap: A performance-agnostic metric for measuring multimodal contributions in vision and language models \& tasks}.
In: \bbtitle{Proceedings of the 61st Annual Meeting of the Association for Computational Linguistics (Volume 1: Long Papers)}.
\bpublisher{Association for Computational Linguistics}, \blocation{???}
(\byear{2023}).
\doiurl{10.18653/v1/2023.acl-long.223} .
\burl{http://dx.doi.org/10.18653/v1/2023.acl-long.223}
\end{bchapter}
\endbibitem

\bibitem[\protect\citeauthoryear{Shapley}{1953}]{shapley1953value}
\begin{barticle}
\bauthor{\bsnm{Shapley}, \binits{L.S.}}:
\batitle{A value for n-person games}.
\bjtitle{Contributions to the Theory of Games}
\bvolume{2}(\bissue{28}),
\bfpage{307}--\blpage{317}
(\byear{1953})
\end{barticle}
\endbibitem

\bibitem[\protect\citeauthoryear{Lundberg and Lee}{2017}]{lundberg2017unified}
\begin{bchapter}
\bauthor{\bsnm{Lundberg}, \binits{S.M.}},
\bauthor{\bsnm{Lee}, \binits{S.-I.}}:
\bctitle{A unified approach to interpreting model predictions}.
In: \bbtitle{Advances in Neural Information Processing Systems},
pp. \bfpage{4765}--\blpage{4774}
(\byear{2017})
\end{bchapter}
\endbibitem

\bibitem[\protect\citeauthoryear{Tsai et~al.}{2022}]{tsai2022faith}
\begin{bchapter}
\bauthor{\bsnm{Tsai}, \binits{C.-P.}},
\bauthor{\bsnm{Yeh}, \binits{C.-K.}},
\bauthor{\bsnm{Ravikumar}, \binits{P.}}:
\bctitle{Faith-shap: The faithful shapley interaction index}.
In: \bbtitle{International Conference on Machine Learning},
pp. \bfpage{21863}--\blpage{21890}
(\byear{2022})
\end{bchapter}
\endbibitem

\bibitem[\protect\citeauthoryear{Zhang et~al.}{2023}]{montecarlo2022}
\begin{botherref}
\oauthor{\bsnm{Zhang}, \binits{J.}},
\oauthor{\bsnm{Sun}, \binits{Q.}},
\oauthor{\bsnm{Liu}, \binits{J.}},
\oauthor{\bsnm{Xiong}, \binits{L.}},
\oauthor{\bsnm{Pei}, \binits{J.}},
\oauthor{\bsnm{Ren}, \binits{K.}}:
Efficient sampling approaches to shapley value approximation.
Proc. ACM Manag. Data
\textbf{1}(1)
(2023)
\doiurl{10.1145/3588728}
\end{botherref}
\endbibitem

\bibitem[\protect\citeauthoryear{Hsieh et~al.}{2022}]{gmdb2022}
\begin{barticle}
\bauthor{\bsnm{Hsieh}, \binits{T.}},
\bauthor{\bsnm{Bar-Haim}, \binits{A.}},
\bauthor{\bsnm{Moosa}, \binits{S.}},
\bauthor{\bsnm{Ehmke}, \binits{N.}},
\bauthor{\bsnm{Gripp}, \binits{K.}},
\bauthor{\bsnm{Pantel}, \binits{J.}},
\bauthor{\bsnm{Danyel}, \binits{M.}},
\bauthor{\bsnm{Mensah}, \binits{M.}},
\bauthor{\bsnm{Horn}, \binits{D.}},
\bauthor{\bsnm{Rosnev}, \binits{S.}},
\bauthor{\bsnm{Fleischer}, \binits{N.}},
\bauthor{\bsnm{Bonini}, \binits{G.}},
\bauthor{\bsnm{Hustinx}, \binits{A.}},
\bauthor{\bsnm{Schmid}, \binits{A.}},
\bauthor{\bsnm{Knaus}, \binits{A.}},
\bauthor{\bsnm{Javanmardi}, \binits{B.}},
\bauthor{\bsnm{Klinkhammer}, \binits{H.}},
\bauthor{\bsnm{Lesmann}, \binits{H.}},
\bauthor{\bsnm{Sivalingam}, \binits{S.}},
\bauthor{\bsnm{Kamphans}, \binits{T.}},
\bauthor{\bsnm{Meiswinkel}, \binits{W.}},
\bauthor{\bsnm{Ebstein}, \binits{F.}},
\bauthor{\bsnm{Kr{\"u}ger}, \binits{E.}},
\bauthor{\bsnm{K{\"u}ry}, \binits{S.}},
\bauthor{\bsnm{B{\'e}zieau}, \binits{S.}},
\bauthor{\bsnm{Schmidt}, \binits{A.}},
\bauthor{\bsnm{Peters}, \binits{S.}},
\bauthor{\bsnm{Engels}, \binits{H.}},
\bauthor{\bsnm{Mangold}, \binits{E.}},
\bauthor{\bsnm{Krei{\ss}}, \binits{M.}},
\bauthor{\bsnm{Cremer}, \binits{K.}},
\bauthor{\bsnm{Perne}, \binits{C.}},
\bauthor{\bsnm{Betz}, \binits{R.}},
\bauthor{\bsnm{Bender}, \binits{T.}},
\bauthor{\bsnm{Grundmann-Hauser}, \binits{K.}},
\bauthor{\bsnm{Haack}, \binits{T.}},
\bauthor{\bsnm{Wagner}, \binits{M.}},
\bauthor{\bsnm{Brunet}, \binits{T.}},
\bauthor{\bsnm{Bentzen}, \binits{H.}},
\bauthor{\bsnm{Averdunk}, \binits{L.}},
\bauthor{\bsnm{Coetzer}, \binits{K.}},
\bauthor{\bsnm{Lyon}, \binits{G.}},
\bauthor{\bsnm{Spielmann}, \binits{M.}},
\bauthor{\bsnm{Schaaf}, \binits{C.}},
\bauthor{\bsnm{Mundlos}, \binits{S.}},
\bauthor{\bsnm{N{\"o}then}, \binits{M.}},
\bauthor{\bsnm{Krawitz}, \binits{P.}}:
\batitle{Gestaltmatcher facilitates rare disease matching using facial phenotype descriptors}.
\bjtitle{Nature Genetics}
\bvolume{54}(\bissue{3}),
\bfpage{349}--\blpage{357}
(\byear{2022})
\doiurl{10.1038/s41588-021-01010-x} .
\bcomment{Publisher Copyright: {\textcopyright} 2022, The Author(s), under exclusive licence to Springer Nature America, Inc.}
\end{barticle}
\endbibitem

\bibitem[\protect\citeauthoryear{Plummer et~al.}{2016}]{flickr30k}
\begin{botherref}
\oauthor{\bsnm{Plummer}, \binits{B.A.}},
\oauthor{\bsnm{Wang}, \binits{L.}},
\oauthor{\bsnm{Cervantes}, \binits{C.M.}},
\oauthor{\bsnm{Caicedo}, \binits{J.C.}},
\oauthor{\bsnm{Hockenmaier}, \binits{J.}},
\oauthor{\bsnm{Lazebnik}, \binits{S.}}:
Flickr30k Entities: Collecting Region-to-Phrase Correspondences for Richer Image-to-Sentence Models
(2016).
\url{https://arxiv.org/abs/1505.04870}
\end{botherref}
\endbibitem

\bibitem[\protect\citeauthoryear{Wu et~al.}{2024}]{gestaltmml2023}
\begin{botherref}
\oauthor{\bsnm{Wu}, \binits{D.}},
\oauthor{\bsnm{Yang}, \binits{J.}},
\oauthor{\bsnm{Liu}, \binits{C.}},
\oauthor{\bsnm{Hsieh}, \binits{T.-C.}},
\oauthor{\bsnm{Marchi}, \binits{E.}},
\oauthor{\bsnm{Blair}, \binits{J.}},
\oauthor{\bsnm{Krawitz}, \binits{P.}},
\oauthor{\bsnm{Weng}, \binits{C.}},
\oauthor{\bsnm{Chung}, \binits{W.}},
\oauthor{\bsnm{Lyon}, \binits{G.J.}},
\oauthor{\bsnm{Krantz}, \binits{I.D.}},
\oauthor{\bsnm{Kalish}, \binits{J.M.}},
\oauthor{\bsnm{Wang}, \binits{K.}}:
GestaltMML: Enhancing Rare Genetic Disease Diagnosis through Multimodal Machine Learning Combining Facial Images and Clinical Texts
(2024).
\url{https://arxiv.org/abs/2312.15320}
\end{botherref}
\endbibitem

\bibitem[\protect\citeauthoryear{Deardorff et~al.}{2020}]{Deardorff2020CdLS}
\begin{botherref}
\oauthor{\bsnm{Deardorff}, \binits{M.A.}},
\oauthor{\bsnm{Noon}, \binits{S.E.}},
\oauthor{\bsnm{Krantz}, \binits{I.D.}}:
Cornelia de Lange Syndrome.
University of Washington, Seattle.
GeneReviews{\textregistered}, updated October 15, 2020
(2020)
\end{botherref}
\endbibitem

\bibitem[\protect\citeauthoryear{Roberts}{2025}]{Roberts2025Noonan}
\begin{botherref}
\oauthor{\bsnm{Roberts}, \binits{A.E.}}:
Noonan Syndrome.
University of Washington, Seattle.
GeneReviews{\textregistered}, updated December 4, 2025
(2025)
\end{botherref}
\endbibitem

\bibitem[\protect\citeauthoryear{Dagli et~al.}{2025}]{Dagli2025Angelman}
\begin{botherref}
\oauthor{\bsnm{Dagli}, \binits{A.I.}},
\oauthor{\bsnm{Mathews}, \binits{J.}},
\oauthor{\bsnm{Williams}, \binits{C.A.}}:
Angelman Syndrome.
University of Washington, Seattle.
GeneReviews{\textregistered}, updated May 1, 2025
(2025)
\end{botherref}
\endbibitem

\bibitem[\protect\citeauthoryear{Selvaraju et~al.}{2017}]{selvaraju2017grad}
\begin{bchapter}
\bauthor{\bsnm{Selvaraju}, \binits{R.R.}},
\bauthor{\bsnm{Cogswell}, \binits{M.}},
\bauthor{\bsnm{Das}, \binits{A.}},
\bauthor{\bsnm{Vedantam}, \binits{R.}},
\bauthor{\bsnm{Parikh}, \binits{D.}},
\bauthor{\bsnm{Batra}, \binits{D.}}:
\bctitle{Grad-cam: Visual explanations from deep networks via gradient-based localization}.
In: \bbtitle{Proceedings of the IEEE International Conference on Computer Vision},
pp. \bfpage{618}--\blpage{626}
(\byear{2017})
\end{bchapter}
\endbibitem

\end{thebibliography}

\end{document}